\pdfoutput=1

\documentclass[11pt]{article}

\usepackage[preprint]{acl}

\usepackage{latexsym}
\usepackage{wrapfig}

\usepackage{amsmath,amsfonts,bm}

\def\eqref#1{equation~\ref{#1}}

\def\1{\bm{1}}

\DeclareMathAlphabet{\mathsfit}{\encodingdefault}{\sfdefault}{m}{sl}
\SetMathAlphabet{\mathsfit}{bold}{\encodingdefault}{\sfdefault}{bx}{n}

\usepackage{times}
\usepackage{color}
\usepackage{bm}
\usepackage{tabularx}
\usepackage{tabularray}
\usepackage{dsfont}
\usepackage[T1]{fontenc}

\usepackage[utf8]{inputenc}

\usepackage{microtype}

\usepackage{inconsolata}

\usepackage{graphicx}
\usepackage{microtype}

\usepackage{hyperref}       %
\usepackage{inconsolata}
\usepackage{algorithm}
\usepackage{wrapfig}
\usepackage{algorithmic}
\usepackage{colortbl}
\usepackage{wrapfig}
\usepackage{xcolor}
\usepackage{inconsolata}
\usepackage{multirow}  
\usepackage{amssymb}   
\usepackage{amsmath}
\usepackage{color}
\usepackage{fontawesome5}
\usepackage{subfigure} 
\usepackage{makecell} 
\usepackage{booktabs} 
\usepackage{enumitem} 
\usepackage{multirow} 

\usepackage{booktabs} 
\usepackage{stfloats}
\usepackage{listings}
\usepackage{cite}
\usepackage{pifont}
\usepackage{tcolorbox}
\usepackage{soul}
\usepackage{enumitem}
\tcbuselibrary{skins, breakable, theorems}
\usepackage{wrapfig}
\usepackage{lscape}
\usepackage{url}

\definecolor{mypink}{rgb}{.99,.91,.95}
\definecolor{myyellow}{rgb}{.99,.94,.82}
\newcommand{\OURS}[0]{NOVA}
\definecolor{Gray}{gray}{0.9}
\definecolor{mygreen}{rgb}{0.0, 0.5, 0.0}
\definecolor{myred}{rgb}{0.8, 0.25, 0.33}
\definecolor{myblue}{rgb}{0.19, 0.55, 0.91}
\definecolor{uclablue}{rgb}{0.15, 0.45, 0.68}
\definecolor{ucladblue}{rgb}{0.0, 0.33, 0.53}
\definecolor{ucladdblue}{rgb}{0.0, 0.23, 0.36}
\definecolor{uclagold}{rgb}{1.0, 0.82, 0.0}
\definecolor{ucladgold}{rgb}{1.0, 0.78, 0.17}
\definecolor{ucladdgold}{rgb}{1.0, 0.72, 0.11}
\definecolor{boxgreen}{rgb}{0.02, 0.66, 0.02}
\definecolor{boxred}{rgb}{0.66, 0.1, 0.1}
\definecolor{boxblue}{rgb}{0.01, 0.01, 0.73}

\usepackage{times}
\usepackage{latexsym}

\usepackage[utf8]{inputenc}
\usepackage[T1]{fontenc}

\usepackage{inconsolata}

\usepackage{algorithm}
\usepackage{algorithmic}

\usepackage{newfloat}
\usepackage{listings}
\floatstyle{ruled}
\newfloat{listing}{tb}{lst}{}
\floatname{listing}{Listing}

\usepackage{graphicx}
\usepackage{amsmath,amssymb,amsthm,mathabx,amsfonts}
\usepackage{bbm}
\usepackage{cleveref}
\usepackage{acronym}
\usepackage{enumitem}
\usepackage{balance}
\usepackage{xspace}
\usepackage{setspace}
\usepackage{url}
\usepackage{amsfonts}
\usepackage{multirow}
\usepackage{booktabs}
\usepackage{tablefootnote}
\usepackage[switch]{lineno}
\usepackage{multicol,lipsum}
\usepackage{tikz}
\usepackage{pgfplots}
\usepackage{pgfplotstable}
\usepackage{arydshln}

\usepackage{CJKutf8}

\pgfplotsset{compat=1.18}
\makeatletter
\DeclareRobustCommand\onedot{\futurelet\@let@token\@onedot}
\def\@onedot{\ifx\@let@token.\else.\null\fi\xspace}

\makeatother

\makeatletter
\newcommand{\thickhline}{%
    \noalign {\ifnum 0=`}\fi \hrule height 1pt
    \futurelet \reserved@a \@xhline
}

\crefname{algorithm}{Alg.}{Algs.}
\Crefname{algocf}{Algorithm}{Algorithms}
\crefname{section}{Sec.}{Secs.}
\Crefname{section}{Section}{Sections}
\crefname{table}{Tab.}{Tabs.}
\Crefname{table}{Table}{Tables}
\crefname{figure}{Fig.}{Fig.}
\Crefname{figure}{Figure}{Figure}
\crefname{appendix}{Appendix}{Appendices}

\acrodef{nlp}[NLP]{natural language processing}
\acrodef{plm}[PLM]{Pre-trained Language Model}
\acrodef{sota}[SOTA]{state-of-the-art}
\acrodef{icl}[ICL]{In-Context Learning}
\acrodef{bbl}[BBL]{BIG-bench Lite}

\usepackage{colortbl}
\definecolor{gblue}{HTML}{4285F4}
\definecolor{gred}{HTML}{DB4437}
\definecolor{ggreen}{HTML}{0F9D58}

\definecolor{mygray}{gray}{.92}
\definecolor{emphypurple}{rgb}{0.302, 0.055, 0.659}

\definecolor{highlightgreen}{HTML}{009901}
\definecolor{highlightred}{HTML}{FD6864}

\title{Aligning Large Language Models to Follow Instructions and Hallucinate Less via Effective Data Filtering}

\author{
\textbf{Shuzheng Si$^{\spadesuit\diamondsuit}$, Haozhe Zhao$^{\heartsuit}$, Gang Chen$^\diamondsuit$, Cheng Gao$^{\spadesuit}$, Yuzhuo Bai$^{\spadesuit}$} \\
\textbf{Zhitong Wang$^\spadesuit$, Kaikai An$^{\heartsuit}$, Kangyang Luo$^{\spadesuit}$, Chen Qian$^{\spadesuit}$} \\ 
\textbf{ 
Fanchao Qi\thanks{\ Corresponding Authors.}$^{\spadesuit}$, Baobao Chang$^{\heartsuit}$,} and \textbf{Maosong Sun\footnotemark[1]$^{\spadesuit}$} \\ 
$^{\spadesuit}$ Tsinghua University \quad $^{\heartsuit}$ Peking University
\quad $^\diamondsuit$ DeepLang AI
}

\begin{document}
\maketitle
\renewcommand{\thefootnote}{\fnsymbol{footnote}}
\renewcommand{\thefootnote}{\arabic{footnote}}
\urlstyle{same}
\definecolor{darkgreen}{RGB}{50,100,0}
\definecolor{darkred}{RGB}{200, 0, 0}
\definecolor{lightred}{RGB}{250, 200, 200}
\definecolor{lightblue}{RGB}{210, 220, 250}
\newcommand{\cmark}{\textcolor{darkgreen}{\ding{51}}} %
\newcommand{\xmark}{\textcolor{darkred}{\ding{55}}} %
\definecolor{tabcolor1}{RGB}{247,225,237} 
\definecolor{tabcolor2}{RGB}{255, 250, 132} 
\definecolor{tabcolor3}{RGB}{204, 232, 207} 
\definecolor{tabcolor4}{RGB}{245, 222, 179} 
\definecolor{tabcolor5}{RGB}{210, 220, 250} 
\definecolor{tabcolor6}{RGB}{237, 237, 237} 

\begin{abstract}
Training LLMs on data containing unfamiliar knowledge during the instruction tuning stage can encourage hallucinations.
To address this challenge, we introduce \textbf{\OURS}, a novel framework designed to identify high-quality data that aligns well with the LLM's learned knowledge to reduce hallucinations. 
\OURS~includes Internal Consistency Probing (ICP) and Semantic Equivalence Identification (SEI) to measure how familiar the LLM is with instruction data. 
Specifically, ICP evaluates the LLM's understanding of the given instruction by calculating the tailored consistency among multiple self-generated responses.
SEI further assesses the familiarity of the LLM with the target response by comparing it to the generated responses, using the proposed semantic clustering and well-designed voting strategy.
Finally, to ensure the quality of selected samples, we introduce an expert-aligned reward model, considering characteristics beyond just familiarity.
By considering data quality and avoiding unfamiliar data, we can utilize the selected data to effectively align LLMs to follow instructions and hallucinate less.
Extensive experiments and analysis show that \OURS~significantly reduces hallucinations and allows LLMs to maintain a strong ability to follow instructions.\footnote{~The data and code will be available at \url{https://github.com/S1s-Z/NOVA}. 
Email: ssz24@mails.tsinghua.edu.cn.}

\end{abstract}
\section{Introduction}
\label{section:introduction}
Alignment is a critical procedure to ensure large language models (LLMs) follow user instructions \citep{ OpenAI2023GPT4TR, yang2024qwen2technicalreport}. 
Despite significant progress in LLM alignment and instruction tuning \citep{ouyang2022training, Bai2022TrainingAH}, state-of-the-art aligned LLMs still generate statements that appear credible but are actually incorrect, referred to as hallucinations \citep{ji-survey, hit-survey}.
Such hallucinations can undermine the trustworthiness of LLMs in real-world applications \citep{si2023spokenwoz, min2023factscore, rawte2023surveyhallucinationlargefoundation, wei2024long}.

\begin{figure}
    \centering
    \includegraphics[width=8cm]{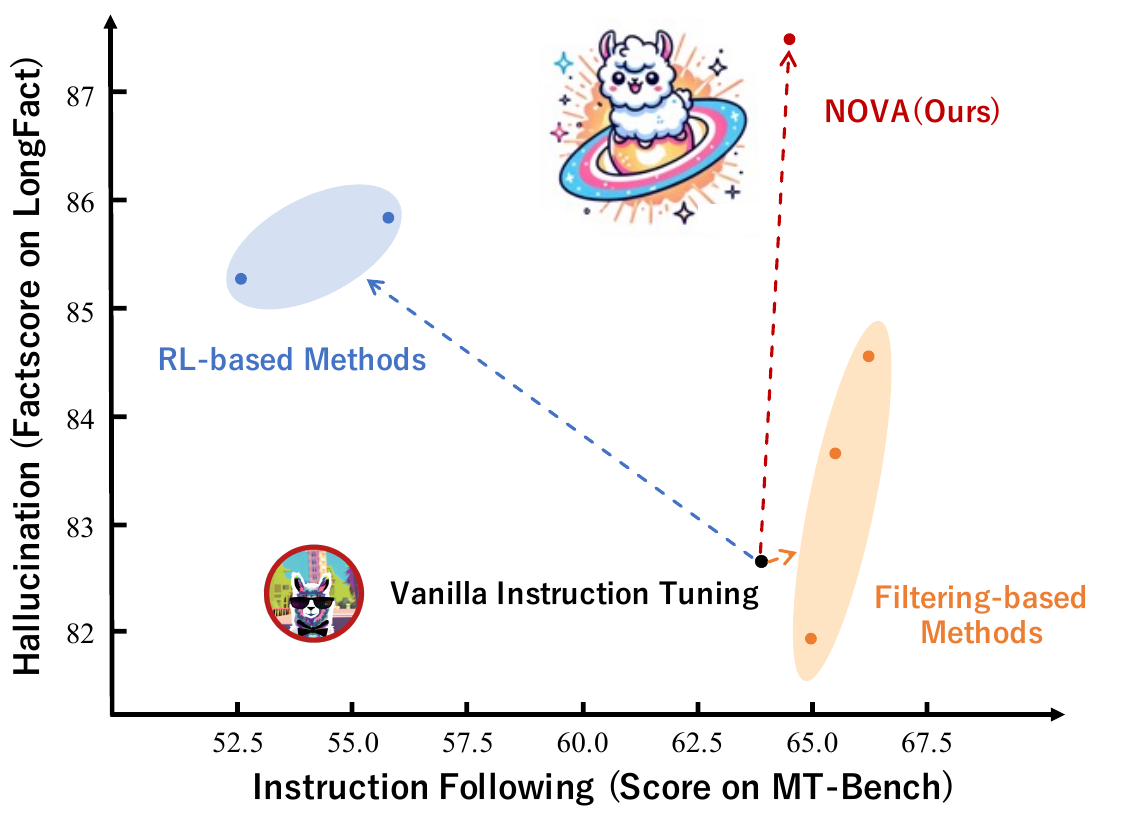}
    \caption{Instruction following ability on MT-Bench vs hallucination on LongFact. \textbf{\OURS} simultaneously aligns LLMs to follow instructions and hallucinate less.}
    \label{fig_example}
\end{figure}

Previous studies \citep{kang2024unfamiliarfinetuningexamplescontrol, gekhman-etal-2024-fine, lin2024flame} indicate that tuning LLMs on instruction data that contains new or unfamiliar knowledge can encourage models to be overconfident and promote hallucinations. 
In other words, once the knowledge in the instruction data has not been learned during the pre-training stage of LLMs, the fine-tuned LLMs tend to produce more errors when generating responses.
Therefore, there is a dilemma in instruction tuning:
On the one hand, the LLMs need to learn to follow user instructions during this stage, which is crucial for user interaction in real-world applications \citep{wang2023how, chen2024alpagasus}; 
On the other hand, using high-quality data (whether manually labeled or generated by other advanced LLMs) for instruction tuning can introduce unfamiliar knowledge to LLMs, thereby encouraging hallucinations \citep{kang2024unfamiliarfinetuningexamplescontrol, lin2024flame}.
Thus, a critical question arises: \textit{\textbf{How can we align LLMs to follow instructions and hallucinate less during the instruction tuning stage?}}

Certain efforts \citep{lin2024flame, zhang-etal-2024-self, tian2024finetuning} apply reinforcement learning (RL) to teach LLMs to hallucinate less after the instruction tuning stage.
For example, \citet{zhang-etal-2024-self} leverages the self-evaluation capability of an LLM and employs GPT-3.5-turbo \citep{chatgpt2022} to create preference data, subsequently aligning the LLM with direct preference optimization (DPO) \citep{dpo}.
However, \citet{lin2024flame} finds that such RL-based methods can weaken the model's ability to follow instructions.
These methods also necessitate additional preference data and API costs from the advanced LLMs, making them inefficient.
Different from RL-based methods, an intuitive strategy to align LLMs to follow instructions and hallucinate less is to filter out the instruction data that contains unfamiliar knowledge for the instruction tuning.
Unfortunately, previous studies \citep{ liu2024what, cao2024instructionmininginstructiondata} solely focus on selecting high-quality data to improve the instruction-following abilities of LLMs.
Even worse, these selected high-quality data may present more unknown knowledge to the LLM and further encourage hallucinations, as these data may contain responses with expert-level knowledge and often delve into advanced levels of detail.

Therefore, we introduce \textbf{\OURS}, which includes \textbf{I\underline{n}ternal C\underline{o}nsistency Probing (ICP)} and \textbf{Semantic Equi\underline{v}alence Identific\underline{a}tion (SEI)}, a framework designed to identify high-quality instruction samples that align well with LLM’s knowledge, thereby aligning the LLM to follow instructions and hallucinate less.
\OURS~initially uses ICP and SEI to measure how well the LLM understands the knowledge in the given instruction and target response. 
For ICP, we prompt the LLM to generate multiple responses to demonstrate what it has learned about a specific instruction during pre-training.
Then we use the internal states produced by the LLM to assess how consistent the generated responses are. 
If the internal states of these responses exhibit greater consistency for the instruction, it indicates that the LLM has internalized the relevant knowledge during pre-training.
For SEI, we first integrate a well-trained model to classify the generated responses that convey the same thing into a semantic cluster.
Next, we employ the designed voting strategy to identify which semantic cluster the target response fits in.
This helps us find out how many generated responses are semantically equivalent to the target response, indicating how well the LLM understands the target response.
If the target response matches well with the largest cluster, it shows the LLM is familiar with its content.
Based on ICP and SEI, we can measure how well the model understands the knowledge in instruction data and avoid training it on unfamiliar data to reduce hallucinations.
Lastly, to ensure the quality of selected samples, we introduce an expert-aligned quality reward model, considering characteristics beyond just familiarity, e.g., the complexity of instructions and the fluency of responses.
By considering data quality and avoiding unfamiliar data, we can use the selected data to effectively align LLMs to follow instructions and hallucinate less.

We conduct extensive experiments to evaluate the effectiveness of \OURS~from both instruction-following and hallucination perspectives.
Experimental results demonstrate that \OURS~significantly reduces hallucinations while maintaining a competitive ability to follow instructions.

\section{Related Work}

\textbf{Hallucinations in LLMs.}
\
Hallucinations occur when the generated content from LLMs seems believable but does not match factual or contextual knowledge \citep{ji-survey, hit-survey, si2025teaching}.
Recent studies \citep{lin2024flame, kang2024unfamiliarfinetuningexamplescontrol, gekhman-etal-2024-fine} attempt to analyze the causes of hallucinations in LLMs and find that tuning LLMs on data containing unseen knowledge can encourage models to be overconfident, leading to hallucinations.
Therefore, recent studies \citep{lin2024flame, zhang-etal-2024-self, tian2024finetuning} attempt to apply RL-based methods to teach LLMs to hallucinate less after the instruction tuning stage.
However, these methods are inefficient because they require additional corpus and API costs for advanced LLMs.
Even worse, such RL-based methods can weaken the instruction-following ability of LLMs \citep{lin2024flame}.
In this paper, instead of introducing the inefficient RL stage, we attempt to directly filter out the unfamiliar data during the instruction tuning stage, aligning LLMs to follow instructions and hallucinate less.

\noindent
\textbf{Data Filtering for Instruction Tuning.}
\
According to \citet{zhou2023lima}, data quality is more important than data quantity in instruction tuning.
Therefore, many works attempt to select high-quality instruction samples to improve the LLMs’ instruction-following abilities.
\citet{chen2023alpagasus, liu2024what} utilize the feedback from well-aligned close-source LLMs to select samples.
\citet{cao2024instructionmininginstructiondata,li-etal-2024-quantity, ge2024clustering, si2024selecting, xia2024less,zhang2024recostexternalknowledgeguided} try to utilize the well-designed metrics (e.g., complexity) based on open-source LLMs to select the samples.
However, these high-quality data always contain expert-level responses and may contain much unfamiliar knowledge to the LLM.
Unlike focusing on data quality, we attempt to identify the samples that align well with LLM's knowledge, thereby allowing the LLM to hallucinate less.

\section{Methodology}
\label{section:method}
\begin{figure*}[t]
    \centering
    \includegraphics[width=0.97\linewidth]{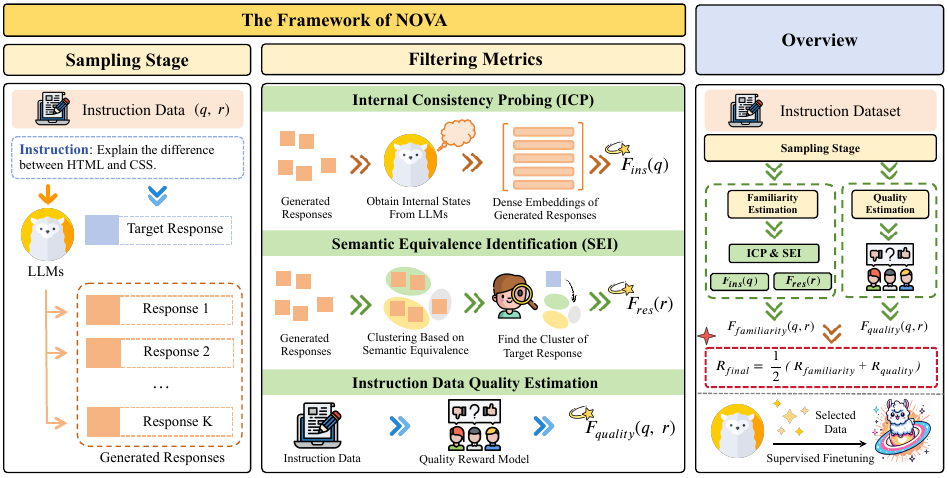}
    \caption{
     The process of \OURS.
     \OURS~identifies and selects high-quality instruction data that aligns well with the LLM’s learned knowledge to reduce hallucination.
     Then it uses selected instruction data for training LLMs.
    }
    \label{figure:model}
\end{figure*}

In this section, we will detail our proposed framework \textbf{\OURS} as shown in Figure \ref{figure:model}.
Previous studies \citep{lin2024flame, kang2024unfamiliarfinetuningexamplescontrol,gekhman-etal-2024-fine} find that tuning LLMs on data containing new or unfamiliar knowledge can encourage models to be overconfident and further lead to hallucinations.
Inspired by this finding, \OURS~aims to filter out the unfamiliar instruction data for the instruction tuning, thereby aligning the LLM to follow instructions and hallucinate less.

\subsection{Internal Consistency Probing}
\label{section:IKP}
To comprehensively measure the LLM's familiarity with instruction data, the first challenge is to evaluate how well the LLM understands the knowledge within the instructions.
Prompting LLMs to generate multiple responses to the same instruction and measuring how consistent those responses are has been proven to be an effective way \citep{wang2023selfconsistency, chen2024inside}.
This is because if LLMs understand the question and are confident in their answers, they will produce similar responses.
A practical way to measure the consistency of free-form responses is to utilize lexical metrics (e.g., Rouge-L) \citep{DBLP:journals/tmlr/LinT024} or sentence-level confidence scores (e.g., perplexity) \citep{ren2023outofdistribution}.
However, these straightforward strategies neglect highly concentrated semantic information within the internal states of LLMs, and thus fail to capture the fine-grained differences between responses.

Hence, we propose \textbf{Internal Consistency Probing (ICP)} to measure the semantic consistency in the dense embedding space.
For an instruction data $s=(q, r)$, $q$ denotes the instruction, and $r$ denotes the target response.
For instruction $q$, we first sample $K$ responses $[r'_1,...,r'_K]$ from a base LLM and apply few-shot demonstrations \citep{lin2024the} to ensure the coherence of generated responses.
For $K$ generated responses, we use the internal states of the last token of each response in the last layer as the final sentence embeddings $E=[e_1,e_2,...,e_K]$, as it effectively captures the sentence semantics \citep{azaria2023the}.
We further utilize differential entropy (DE) to assess the semantic consistency in continuous embedding space, which is the extension of discrete Shannon entropy:
\begin{align}
{\rm DE}(X)=-{\rm \int}_xf(x)~{\rm log}(f(x))dx.
\end{align}

We process and treat sentence embeddings $E$ as a multivariate Gaussian Distribution $E\sim N(\mu, \Sigma)$.
Then, the differential entropy can be expressed as:
\begin{align}
\label{eq:de}
{\rm DE}(E) &= \frac{1}{2}{\rm log}((2\pi e)^d{\rm det} (\Sigma)),
\end{align}
where ${\rm det}(\Sigma)$ represents the determinant of the covariance matrix $\Sigma$, $d$ is the dimension of the sentence embedding, and $e$ is the natural constant.
$\Sigma$ denotes the covariance matrix that captures the relationship between $K$ different sentence embeddings, which takes the form:
\begin{align}
\Sigma = \frac{1}{K-1} \sum_{i=1}^{K}(e_i-\mu)(e_i-\mu)^T.
\end{align}

Finally, we measure semantic consistency using ${\rm DE}(E)$, term as \bm{$F_{ins}(q)$} for a given instruction $q$ in data $s$.
Also, ${\rm DE}(E)$ in Eq.(\ref{eq:de}) simplifies to:
{
\small
\begin{align}
\label{eq:de_final}
F_{ins}(q) = \frac{1}{2}{\rm log}{\rm det}(\Sigma) + \frac{d}{2}({\rm log}2\pi+1) = \frac{1}{2}\sum_{i=1}^{d}\lambda_i + G,
\end{align}
}

\noindent
where $\lambda_i$ denotes the $i$-th eigenvalue of the covariance matrix $\Sigma$, which can be easily calculated by singular value decomposition.
$G$ is a constant.

If the LLM is familiar with the given instruction, the sentence embeddings of generated responses will be highly correlated and the value of $F_{ins}(q)$ will be close to $G$.
On the contrary, when the LLM is indecisive, the model will generate multiple responses with different meanings leading to a significant value of $F_{ins}(q)$. 
In this way, we can exploit the dense semantic information to effectively measure the LLM’s familiarity with the instruction.

\subsection{Semantic Equivalence Identification}
\label{section:SEI}
Another challenge is to estimate the knowledge in the target response and measure the LLM's familiarity with it, since the target response can contain expert-level and unfamiliar knowledge for the LLM.
Training LLMs on such data can encourage hallucinations.
Therefore, we propose \textbf{Semantic Equivalence Identification (SEI)} to measure the LLM’s familiarity with the target response by calculating how many generated responses are semantically equivalent to the target response.
If the target response and more generated responses convey the same meaning, it indicates that the LLM is more familiar with it, thereby training the LLM on this target response will reduce hallucinations.

As the target response is manually labeled or derived from advanced LLMs (e.g., GPT-4) instead of generated by the LLM itself, the internal states of the LLM cannot effectively represent the target response.
Thus, unlike utilizing internal states as the proposed ICP, we calculate LLM's familiarity with target responses using the proposed semantic clustering strategy.
In detail, we first cluster the generated responses that convey the same thing into a semantic cluster.
This is because these responses are often free-form, and multiple generated responses can have the same meaning in different ways.
Therefore, we employ an off-the-shelf natural language inference (NLI) model to cluster these responses.
NLI models are trained to infer the logical entailment between an arbitrary pair of sentences.
Thus, NLI models are well-suited to identify semantic equivalence, as two generated responses mean the same thing if you can entail (i.e. logically imply) each from the other \citep{kuhn2023semantic, jung-etal-2024-impossible}.
In this way, we can use an NLI model to consider two responses that can be entailed from each other as semantically equivalent responses.
Specifically, we test each pair $(r'_i, r'_j)$ of $i$-th and $j$-th generated responses as:
\begin{align}
\small
\begin{split}
    F_{\textit{equivalent}}(r'_i, r'_j) = \mathbb I \Bigl\{ & L_{\textit{NLI}}(r'_i \Rightarrow r'_j) = L_{\textit{entailment}} \,\, \land \\
    & L_{\textit{NLI}}(r'_j \Rightarrow r'_i) =L_{\textit{entailment}} \Bigr\},
\end{split}
\end{align}
where $L_{NLI}$ represents the predictions of the NLI model, $L_{entailment}$ means the label of entailment relation.
$\mathbb I$ is the indicator function.

In this way, we can identify the semantic equivalence of each pair of generated responses and then cluster these generated responses $[r'_1,...,r'_K]$ into $M$ different semantic clusters $[c_1,...,c_M]$, where $m$-th semantic cluster $c_m$ contains $k_m$ generated responses.
Each semantic cluster $c$ is a set of generated responses that convey the same thing.
We further apply the NLI model to determine which semantic cluster the target response $r$ fits in.
Specifically, we use the model to test the target response $r$ and each generated response $r'_i \in [r'_1,...,r'_K]$: 
\begin{align}
\small
\begin{split}
    F_{\textit{equivalent}}(r, r'_i) = \mathbb I \Bigl\{ & L_{\textit{NLI}}(r \Rightarrow r'_i) = L_{\textit{entailment}} \,\, \land \\
    & L_{\textit{NLI}}(r'_i \Rightarrow r) =L_{\textit{entailment}} \Bigr\}.
\end{split}
\end{align}

Using this method, we can determine how many generated responses in a semantic cluster are semantically equivalent to the target response $r$.
For semantic clusters $[c_1,...,c_M]$, the counts of such generated responses are $[k'_1,k'_2,...,k'_M]$.
We use the votes in each semantic cluster to decide which cluster the target response belongs to:
\begin{align}
{\rm Index}(c_{target}) = \mathop{\arg\max}([\frac{k'_1}{k_1}, \frac{k'_2}{k_2},..., \frac{k'_M}{k_M}]).
\end{align}

We calculate the ratio of the number of responses $k_{target}$ in the target cluster $c_{target}$ to the total number of generated responses as \bm{$F_{res}(r)$}:
\begin{align}
\label{eq:res}
F_{res}(r) = \frac{k_{target}}{\sum_{m=1}^M k_m}.
\end{align}

According to Eq.(\ref{eq:res}), when the LLM is familiar with the knowledge within the target response $r$, most of the generated responses will have the same meaning as target response $r$, thus the value of $F_{res}(r)$ will be close to 1.
On the contrary, if the target response contains unseen knowledge, i.e., none of the generated responses have the same meaning as it, the value of $F_{res}(r)$ will be close to 0. 
To this end, we can effectively measure the LLM’s familiarity with the target response.


\subsection{Ranking, Selecting, and Training}
To comprehensively estimate the knowledge and consider both the LLM’s familiarity with the instruction and the target response, we calculate the ratio between $F_{ins}(q)$ and $F_{res}(r)$ for an instruction data $(q,r)$ as the final score:
\begin{align}
F_{familiarity}(q,r) = \frac{F_{res}(r)}{F_{ins}(q)}.
\end{align}

This score effectively measures how well the LLM understands the knowledge in instruction data. 
High $F_{familiarity}$ values indicate that the knowledge in the data aligns well with the LLM, as they show that the generated responses are very consistent for a given instruction (i.e., low $F_{ins}(q)$ values) and the generated responses are very semantically similar to the target response (i.e., high $F_{res} (r)$ values).
Based on the principle of filtering unfamiliar instruction data, the data with high $F_{familiarity}$ should be selected to train the LLM.

However, our early experiments observed that selecting instruction data solely based on the LLM's familiarity $F_{familiarity}$ significantly reduces hallucinations but hinders the model's ability to follow instructions.
This is because considering only familiarity ignores other important characteristics of instruction data, e.g., the complexity of the instruction and the fluency of the response.
Therefore, we further introduce an expert-aligned quality reward model to measure the data quality.
We use an expert-labeled preference dataset \citep{DBLP:conf/icde/LiuTZZMZ0HZZMZY24} which contains 3,751 instruction data to train a reward model (more details are shown in Appendix \ref{appendix:id}).
To take both familiarity $F_{familiarity}(q,r)$ and quality $F_{quality}(q,r)$ into consideration, we define the mixed rank $R^{(i)}_{final}$ for $i$-th data as the average of the two ranks corresponding to the two metrics:
\begin{align}
\small
\label{eq:final_r}
R^{(i)}_{final} = \frac{1}{2}(R^{(i)}_{familiarity} + R^{(i)}_{quality}),
\end{align}
where $R^{(i)}_{familiarity}$ and $R^{(i)}_{quality}$ refer to the ranks of the $i$-th data point in the degree of familiarity and quality.
In this way, we can effectively consider data quality and avoid unfamiliar data.

Finally, we rank all the instruction data with their corresponding mixed rank $R_{final}$ to select the top-ranked data, e.g., selecting the top 5\% data to apply the supervised finetuning on the LLM.
Based on the proposed \OURS, we can use the suitable data to effectively align LLMs to follow instructions and hallucinate less during the instruction tuning stage.

\label{section:ranking}

\section{Experiment}
\label{section:experiment}

In this section, we conduct experiments and provide analyses to justify the effectiveness of \OURS.

\subsection{Setup}

\textbf{Instruction Dataset.}
\
We conduct instruction tuning with two different instruction datasets.
\textbf{Alpaca} \citep{alpaca} contains 52,002 samples that are created by employing Text-Davinci-003 model \citep{ouyang2022training} and Self-instruct framework \citep{wang-etal-2023-self-instruct}.
\textbf{Alpaca-GPT4} \citep{peng2023instructiontuninggpt4} further employs more powerful GPT-4 \citep{GPT-4} to get high-quality instruction data.

\begin{table*}[t]
\scriptsize
\centering  
\resizebox{0.97\textwidth}{!}{
\begin{tabular}{lccccccccccccccc}
\toprule
\multirow{2}{*}{\textbf{Model}} & \multicolumn{3}{c}{\textbf{BioGEN$^\dag$ }} & \multicolumn{3}{c}{\textbf{LongFact$^\dag$}} & \multicolumn{5}{c}{\textbf{FollowRAG - Faithfulness$^\ddag$}} \\
\cmidrule(lr){2-4} \cmidrule(lr){5-7} \cmidrule(lr){8-12} 
& \textbf{FactScore} & \textbf{Respond} & \textbf{Facts}  & \textbf{Objects} & \textbf{Concepts} & \textbf{Avg.} & \textbf{NaturalQA} & \textbf{TriviaQA} & \textbf{HotpotQA} & \textbf{WebQSP} & \textbf{Avg.} \\
\midrule
\rowcolor{mygray} \multicolumn{12}{c}{\cellcolor{myyellow} \textbf{Alpaca}} \\
Vanilla - 100\% & 42.4 & 100.0 & 17.1 & 85.8 & 80.3 & 83.1 & 40.5 & 53.5 & 16.0 & 49.5 & 39.9 \\
FLAME-DPO$^{fact}$ & 47.2 & 100.0 & 15.6 & 88.3 & 81.2 & 84.8 & 43.5 & 57.0 & 17.5 & 52.0 & 42.5 \\
SELF-EVAL & 48.3 & 100.0 & 16.9 & 87.8 & 81.0 & 84.4 & 43.0 & 58.0 & 16.5 & 52.5 & 42.5 \\
\midrule
IFD - 5\% & 48.1 & 100.0 & \textbf{21.0} & 87.2 & 80.5 & 83.9 & 41.5 & 57.0 & 15.5 & 51.5 & 41.4 \\
CaR - 5\% & 47.9 & 100.0 & 16.2 & 86.6 & 79.1 & 82.9 & 42.5 & 58.0 & 16.5 & 51.0 & 42.0 \\
Nuggets - 5\% & 48.2 & 100.0 & 18.3 & 88.6 & 81.2 & 84.9 & 42.5 & 56.0 & 16.5 & 51.0 & 41.5 \\
\rowcolor{blue!5} \textbf{\OURS} - 5\% & \textbf{50.3} & 100.0 & 17.9 & \textbf{92.4} & \textbf{82.7} & \textbf{87.6} & \textbf{46.5} & \textbf{60.0} & \textbf{19.0} & \textbf{53.5} & \textbf{44.8} \\
\hdashline[2pt/3pt]
\rowcolor{blue!5}  $\Delta$ compared to Vanilla - 100\%  & \textcolor[rgb]{0.7,0,0}{+7.9} & - & \textcolor[rgb]{0.7,0,0}{+0.8} & \textcolor[rgb]{0.7,0,0}{+6.6} & \textcolor[rgb]{0.7,0,0}{+2.4} & \textcolor[rgb]{0.7,0,0}{+4.5} & \textcolor[rgb]{0.7,0,0}{+6.0} & \textcolor[rgb]{0.7,0,0}{+6.5} & \textcolor[rgb]{0.7,0,0}{+3.0} & \textcolor[rgb]{0.7,0,0}{+4.0} & \textcolor[rgb]{0.7,0,0}{+4.9} \\
\midrule
IFD - 10\% & 43.2 & 100.0 & 20.5 & 86.3 & 79.2 & 82.8 & 40.5 & 60.0 & 17.5 & 53.5 & 42.9 \\
CaR - 10\% & 45.2 & 100.0 & 24.3 & 87.1 & 81.3 & 84.2 & 44.0 & 59.5 & 18.0 & 48.5 & 42.5 \\
Nuggets - 10\% & 45.8 & 100.0 & \textbf{27.1} & 86.7 & 80.4 & 83.6 & 43.0 & 58.5 & 17.0 & 52.5 & 42.8 \\
\rowcolor{blue!5} \textbf{\OURS} - 10\% & \textbf{46.8} & 100.0 & 18.4 & \textbf{89.1} & \textbf{81.6} & \textbf{85.4} & \textbf{46.0} & \textbf{63.0} & \textbf{20.0} & \textbf{59.0} & \textbf{47.0} \\
\hdashline[2pt/3pt]
\rowcolor{blue!5}  $\Delta$ compared to Vanilla - 100\%  & \textcolor[rgb]{0.7,0,0}{+4.4} & - & \textcolor[rgb]{0.7,0,0}{+1.3} & \textcolor[rgb]{0.7,0,0}{+3.3} & \textcolor[rgb]{0.7,0,0}{+1.3} & \textcolor[rgb]{0.7,0,0}{+2.3} & \textcolor[rgb]{0.7,0,0}{+5.5} & \textcolor[rgb]{0.7,0,0}{+9.5} & \textcolor[rgb]{0.7,0,0}{+4.0} & \textcolor[rgb]{0.7,0,0}{+9.5} & \textcolor[rgb]{0.7,0,0}{+7.1} \\
\midrule
IFD - 15\% & 42.2 & 100.0 & 19.4 & 84.7 & 80.7 & 82.7 & 43.5 & 63.0 & 23.0 & 50.0 & 44.9 \\
CaR - 15\% & 43.9 & 100.0 & 20.9 & 86.4 & 78.0 & 82.2 & 45.5 & 61.5 & 22.0 & 48.0 & 44.3 \\
Nuggets - 15\% & 44.3 & 100.0 & \textbf{23.4} & 86.5 & 80.1 & 83.3 & 45.0 & 62.5 & 21.0 & 49.0 & 44.4 \\
\rowcolor{blue!5} \textbf{\OURS} - 15\% & \textbf{45.9} & 100.0 & 18.7 & \textbf{88.1} & \textbf{82.1} & \textbf{85.1} & \textbf{48.5} & \textbf{68.0} & \textbf{25.0} & \textbf{52.0} & \textbf{48.4} \\
\hdashline[2pt/3pt]
\rowcolor{blue!5}  $\Delta$ compared to Vanilla - 100\%  & \textcolor[rgb]{0.7,0,0}{+3.5} & - & \textcolor[rgb]{0.7,0,0}{+1.6}  & \textcolor[rgb]{0.7,0,0}{+2.3} & \textcolor[rgb]{0.7,0,0}{+1.8} & \textcolor[rgb]{0.7,0,0}{+2.0} & \textcolor[rgb]{0.7,0,0}{+8.0} & \textcolor[rgb]{0.7,0,0}{+14.5} & \textcolor[rgb]{0.7,0,0}{+9.0} & \textcolor[rgb]{0.7,0,0}{+2.5} & \textcolor[rgb]{0.7,0,0}{+8.5} \\
\midrule
\multicolumn{16}{c}{\cellcolor{myyellow} \textbf{Alpaca - GPT4}} \\
Vanilla - 100\% & 41.9 & 100.0 & 32.0 & 84.7 & 80.4 & 82.6 & 39.5 & 49.5 & 14.5 & 49.0 & 38.1 \\
FLAME-DPO$^{fact}$ & 46.3 & 100.0 & 27.6 & 87.3 & 84.1 & 85.7 & 42.0 & 55.5 & 16.5 & 52.0 & 41.5 \\
SELF-EVAL & 47.2 & 100.0 & 31.6 & 86.7 & 83.7 & 85.2 & 43.5 & 59.0 & 15.5 & 51.5 & 42.4 \\
\midrule
IFD - 5\% & 46.7 & 100.0 & 39.2 & 84.4 & 79.6 & 82.0 & 42.5 & 58.0 & 16.5 & 52.0 & 42.3 \\
CaR - 5\% & 46.9 & 100.0 & 41.1 & 86.2 & 81.1 & 83.7 & 43.5 & 57.5 & 17.0 & 51.5 & 42.4 \\
Nuggets - 5\% & 47.2 & 100.0 & \textbf{42.3} & 87.0 & 82.3 & 84.7 & 41.0 & 56.0 & 17.0 & 52.0 & 41.5 \\
\rowcolor{blue!5} \textbf{\OURS} - 5\% & \textbf{50.5} & 100.0 & 33.8 & \textbf{90.1} & \textbf{85.2} & \textbf{87.7} & \textbf{45.0} & \textbf{62.0} & \textbf{20.5} & \textbf{53.5} & \textbf{45.3} \\
\hdashline[2pt/3pt]
\rowcolor{blue!5}  $\Delta$ compared to Vanilla - 100\%  & \textcolor[rgb]{0.7,0,0}{+8.6} & - & \textcolor[rgb]{0.7,0,0}{+1.8} & \textcolor[rgb]{0.7,0,0}{+5.4} & \textcolor[rgb]{0.7,0,0}{+4.8} & \textcolor[rgb]{0.7,0,0}{+5.1} & \textcolor[rgb]{0.7,0,0}{+5.5} & \textcolor[rgb]{0.7,0,0}{+12.5} & \textcolor[rgb]{0.7,0,0}{+6.0} & \textcolor[rgb]{0.7,0,0}{+4.5} & \textcolor[rgb]{0.7,0,0}{+7.2} \\
\midrule
IFD - 10\% & 43.6 & 100.0 & \textbf{39.2} & 86.5 & 77.8 & 82.2 & 40.5 & 56.0 & 16.0 & 49.5 & 40.5 \\
CaR - 10\% & 45.9 & 100.0 & 38.0 & 87.1 & 78.3 & 82.7 & 43.0 & 55.0 & 15.5 & 48.0 & 40.4 \\
Nuggets - 10\% & 46.8 & 100.0 & 35.7 & 88.2 & 80.1 & 84.2 & 41.5 & 54.5 & 16.5 & 50.0 & 40.6 \\
\rowcolor{blue!5} \textbf{\OURS} - 10\% & \textbf{48.1} & 100.0 & 32.3 & \textbf{90.6} & \textbf{81.8} & \textbf{86.2} & \textbf{44.5} & \textbf{59.0} & \textbf{18.0} & \textbf{51.0} & \textbf{43.1} \\
\hdashline[2pt/3pt]
\rowcolor{blue!5}  $\Delta$ compared to Vanilla - 100\%  & \textcolor[rgb]{0.7,0,0}{+6.2} & - & \textcolor[rgb]{0.7,0,0}{+0.3}  & \textcolor[rgb]{0.7,0,0}{+5.9} & \textcolor[rgb]{0.7,0,0}{+1.4} & \textcolor[rgb]{0.7,0,0}{+3.6} & \textcolor[rgb]{0.7,0,0}{+5.0} & \textcolor[rgb]{0.7,0,0}{+9.5} & \textcolor[rgb]{0.7,0,0}{+3.5} & \textcolor[rgb]{0.7,0,0}{+2.0} & \textcolor[rgb]{0.7,0,0}{+5.0} \\
\midrule
IFD - 15\% & 42.9 & 100.0 & 32.2 & 85.2 & 80.3 & 82.8 & 46.0 & 54.5 & 15.0 & 52.0 & 41.9 \\
CaR - 15\% & 44.6 & 100.0 & 33.6 & 85.8 & 81.5 & 83.7 & 43.5 & 55.0 & 18.0 & 53.5 & 42.5 \\
Nuggets - 15\% & 44.8 & 100.0 & \textbf{34.5} & 86.1 & 80.7 & 83.4 & 45.0 & 52.0 & 16.0 & 53.0 & 41.5 \\
\rowcolor{blue!5} \textbf{\OURS} - 15\% & \textbf{46.9} & 100.0 & 32.1 & \textbf{88.0} & \textbf{82.5} & \textbf{85.3} & \textbf{49.5} & \textbf{56.5} & \textbf{18.5} & \textbf{55.0} & \textbf{44.9} \\
\hdashline[2pt/3pt]
\rowcolor{blue!5}  $\Delta$ compared to Vanilla - 100\%  & \textcolor[rgb]{0.7,0,0}{+5.0} & - & \textcolor[rgb]{0.7,0,0}{+0.1}  & \textcolor[rgb]{0.7,0,0}{+3.3} & \textcolor[rgb]{0.7,0,0}{+2.1} & \textcolor[rgb]{0.7,0,0}{+2.7} & \textcolor[rgb]{0.7,0,0}{+10.0} & \textcolor[rgb]{0.7,0,0}{+7.0} & \textcolor[rgb]{0.7,0,0}{+4.0} & \textcolor[rgb]{0.7,0,0}{+6.0} & \textcolor[rgb]{0.7,0,0}{+6.8} \\
\bottomrule
\end{tabular}
}
\caption{Results on three hallucination benchmarks. 
$\dag$ indicates the factuality hallucination benchmark. $\ddag$ indicates the faithfulness hallucination benchmark. 
We conduct the experiments based on LLaMA-3-8B.}
\label{tb:main-hall}
\end{table*}

\noindent
\textbf{Evaluation.}
\
To evaluate our method comprehensively, we select widely adopted benchmarks for the targeted abilities.
(1) Factuality hallucination benchmark: BioGEN \citep{min2023factscore} and LongFact \citep{wei2024longformfactualitylargelanguage};
(2) Faithfulness hallucination benchmark: FollowRAG-Faithfulness \citep{dong2024generalinstructionfollowingalignmentretrievalaugmented}, including 4 different QA datasets;
(3) Instruction-following benchmark: MT-Bench \citep{zheng2023judgingllmasajudgemtbenchchatbot} and FollowRAG-Instruction. 
Comprehensive descriptions of tasks, datasets, and evaluation metrics are detailed in Appendix \ref{appendix:eva}.

\noindent
\textbf{Baselines.}
\
We compare several strong baselines, including (1) Vanilla Instruction Tuning: \textbf{Vanilla - 100\%} fine-tunes the model on the whole instruction dataset; 
(2) Instruction Data Filtering Methods: 
\textbf{IFD} \citep{li-etal-2024-quantity} proposes instruction-following difficulty to select a subset of instruction data.
\textbf{CaR} \citep{ge2024clustering} simultaneously considers the data quality and diversity by introducing two scoring methods.
\textbf{Nuggets} \citep{li2024shot}
focuses on selecting high-quality data by identifying samples that notably boost the performance of different tasks after being learned as one-shot instances;
(3) RL-based Methods: \textbf{FLAME-DPO$^{\rm fact}$} \citep{lin2024flame} introduces atomic fact decomposition and retrieval augmented claim verification to construct preference data and apply DPO.
\textbf{SELF-EVAL} \citep{zhang-etal-2024-self} leverages the self-evaluation capability of LLMs and employs GPT-3.5 to create preference data, aligning the LLM with DPO.
We apply these RL-based methods after tuning LLMs on the whole instruction dataset.

\noindent
\textbf{Implementation Details.}
\
Our main experiments are conducted on LLaMA-3-8B and LLaMA-3-70B \citep{llama3}.
More implementation details are shown in Appendix \ref{appendix:id}, e.g., the training of quality reward model and hyperparameters.

\begin{table}[h]
\scriptsize	
\centering
\resizebox{0.91\linewidth}{!}{
\begin{tabular}{lcc}
\toprule
\textbf{Model} & \textbf{MT-Bench} & \textbf{FollowRAG-Intruction}\\
\midrule
\multicolumn{3}{c}{\cellcolor{myyellow} \textbf{Alpaca}}\\
Vanilla - 100\% & 51.9 & 38.7  \\
FLAME-DPO$^{fact}$ & 46.7 & 39.2 \\
SELF-EVAL & 48.3 & 38.5 \\
\midrule
IFD - 5\% & 60.1 & 39.6 \\
CaR - 5\% & 56.6 & \textbf{41.4} \\
Nuggets - 5\% & 60.0 & 40.6 \\
\rowcolor{blue!5} \textbf{\OURS \ - 5\%} & \textbf{60.5} & 39.1 \\
\hdashline[2pt/3pt]
\rowcolor{blue!5} $\Delta$ compared to Vanilla - 100\% & \textcolor[rgb]{0.7,0,0}{+8.6} & \textcolor[rgb]{0.7,0,0}{+0.4} \\
\midrule
IFD - 10\% & 57.2 & 40.4 \\
CaR - 10\% & \textbf{58.3} & \textbf{42.3} \\
Nuggets - 10\% & 58.2 & 41.1 \\
\rowcolor{blue!5} \textbf{\OURS\ - 10\%} & 56.6 & 38.8 \\
\hdashline[2pt/3pt]
\rowcolor{blue!5} $\Delta$ compared to Vanilla - 100\% & \textcolor[rgb]{0.7,0,0}{+4.7} & \textcolor[rgb]{0.7,0,0}{+0.1} \\
\midrule
IFD - 15\% & 56.0 & 40.2 \\
CaR - 15\% & \textbf{57.4} & \textbf{41.0} \\
Nuggets - 15\% & 57.0 & 40.6 \\
\rowcolor{blue!5} \textbf{\OURS\ - 15\%} & 57.2 & 40.1 \\
\hdashline[2pt/3pt]
\rowcolor{blue!5} $\Delta$ compared to Vanilla - 100\% & \textcolor[rgb]{0.7,0,0}{+5.3} & \textcolor[rgb]{0.7,0,0}{+1.4} \\
\midrule
\multicolumn{3}{c}{\cellcolor{myyellow} \textbf{Alpaca - GPT4}}\\
Vanilla - 100\% & 64.3 & 36.9 \\
FLAME-DPO$^{fact}$ & 56.2 & 37.2 \\
SELF-EVAL & 53.1 & 36.5 \\
\midrule
IFD - 5\% & 65.0 & 37.0 \\
CaR - 5\% & 65.4 & 38.0 \\
Nuggets - 5\% & \textbf{66.2} & \textbf{38.5} \\
\rowcolor{blue!5} \textbf{\OURS - 5\%} & 64.6 & 37.8 \\
\hdashline[2pt/3pt]
\rowcolor{blue!5} $\Delta$ compared to Vanilla - 100\% & \textcolor[rgb]{0.7,0,0}{+0.3} & \textcolor[rgb]{0.7,0,0}{+0.9} \\
\midrule
IFD - 10\% & 65.0 & 37.8 \\
CaR - 10\% & 65.8 & 38.0 \\
Nuggets - 10\% & \textbf{67.5} & 38.0 \\
\rowcolor{blue!5} \textbf{\OURS\ - 10\%} & 64.6 & \textbf{39.1} \\
\hdashline[2pt/3pt]
\rowcolor{blue!5} $\Delta$ compared to Vanilla - 100\% & \textcolor[rgb]{0.7,0,0}{+0.3} & \textcolor[rgb]{0.7,0,0}{+2.1} \\
\midrule
IFD - 15\% & 62.3 & 37.9 \\
CaR - 15\% & 61.1 & \textbf{38.1} \\
Nuggets - 15\% & \textbf{66.5} & 38.0 \\
\rowcolor{blue!5} \textbf{\OURS\ - 15\%} & 64.5 & 37.5 \\
\hdashline[2pt/3pt]
\rowcolor{blue!5} $\Delta$ compared to Vanilla - 100\% & \textcolor[rgb]{0.7,0,0}{+0.2} & \textcolor[rgb]{0.7,0,0}{+0.5} \\
\bottomrule
\end{tabular}}
\caption{Results on two instruction-following benchmarks implemented on LLaMA-3-8B.}
\label{tb:if} 
\end{table}


\subsection{Main Results}
\textbf{\OURS~Significantly Reduces Hallucinations.}
\
As shown in Table \ref{tb:main-hall}, \OURS~shows consistent and significant improvements on three hallucination benchmarks measuring factuality and faithfulness.
Compared to indiscriminately using the whole instruction dataset (i.e., Vanilla - 100\%), using samples selected by \OURS~to train LLMs can improve \textbf{3.5-8.6\%} on BioGEN, \textbf{2.0-5.1\%} on LongFact, and \textbf{4.9-8.5\%} on FollowRAG-Faithfulness.
This is because \OURS~effectively filters out the unfamiliar instruction data and avoids training LLMs on these data thereby reducing the hallucinations.
Compared to instruction data filtering methods that focus on data quality, like IFD, our method consistently improves the performance across different selected sample ratios (5-15\%) on three benchmarks.
Meanwhile, these data selected by quality-focused methods may present unfamiliar knowledge to the LLM and encourage hallucinations on LongFact.
On the contrary, \OURS~aims to identify the samples that align well with LLM’s knowledge, helping the LLM to hallucinate less.
\OURS~also achieves better performance than RL-based methods without introducing additional preference data.
These findings underline the effectiveness of our method in aligning LLMs to hallucinate less.

\noindent
\textbf{\OURS~Maintains a Good Balance between Following Instructions and Reducing Hallucinations.}
\
As shown in Table \ref{tb:if}, \OURS~achieves a better instruction-following ability compared to vanilla tuning methods, especially when the LLM is trained on Alpaca.
It shows that \OURS~can effectively align LLMs to follow instructions.
In some cases, our method surpasses data filtering methods that enhance instruction-following ability, demonstrating its effectiveness in identifying suitable data for LLMs.
Unlike RL-based methods that weaken the model's instruction-following ability, our method shows superior instruction-following ability while greatly reducing hallucinations.

\begin{figure}
    \centering
    \includegraphics[width=7cm]{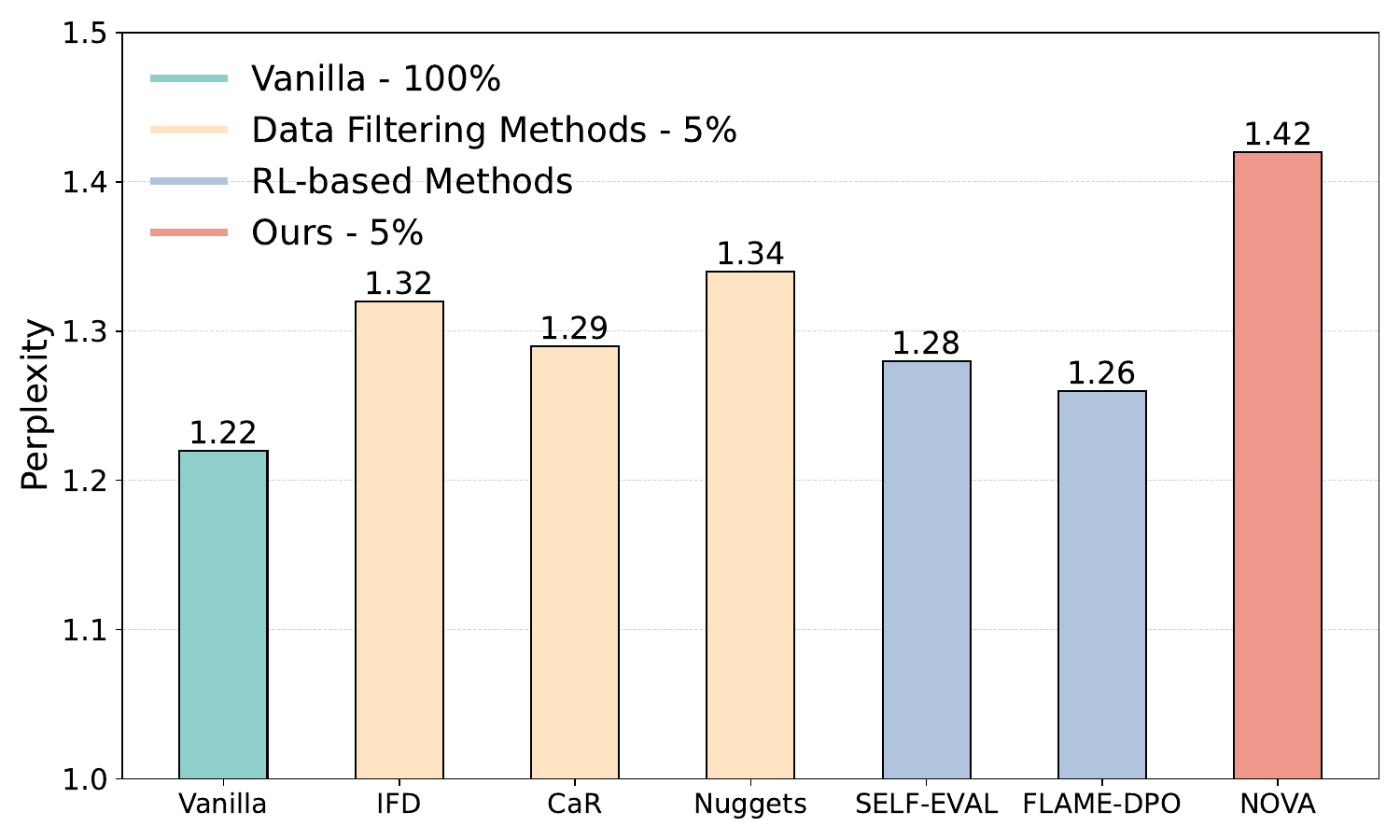}
    \caption{Average perplexity score of 15 samples with the lowest scores for each model from LongFact-Objects. Models are trained on Alpaca-GPT4.}
    \label{fig_ppl}
\end{figure}

\noindent
\textbf{\OURS~Mitigates Overconfidence Phenomenon.}
\
We select 15 samples with the lowest scores for each model from LongFact-Objects and calculate its average perplexity on these samples.
We find that \OURS~generates a high perplexity score (i.e., low sentence-level confidence score) on these bad cases as shown in Figure\ref{fig_ppl}, showing that \OURS~mitigates overconfidence in these false statements.

\begin{table}
\scriptsize	
\centering
\resizebox{0.7\linewidth}{!}{
\begin{tabular}{lcc}
\toprule
\textbf{Model} & \textbf{BioGEN} & \textbf{MT-Bench}\\
\midrule
\rowcolor{blue!5} \textbf{\OURS \ - 5\% - 70B} & 60.9 &  74.3 \\
-w/o. Data Filtering & {53.7} & {73.2}  \\
\hdashline[2pt/3pt]
\rowcolor{blue!5} \textbf{\OURS \ - 5\% - 8B} & 50.5 & 64.6  \\
-w/o. Data Filtering & {41.9} & {64.3}  \\
-w/o. ICP & {47.6} &  {64.1} \\
-w/o. SEI & {48.3} & {63.8} \\
-w/o. Quality RM & {55.6} & {48.6} \\
-w/o. ICP \& SEI & {43.7} & {65.2} \\
\bottomrule
\end{tabular}}
\caption{Results of ablation and scalability study. We report FactScore results on BioGEN. Models are trained on Alpaca-GPT4. RM represents the reward model.}
\label{tb:ab} 
\end{table}

\subsection{Analysis}
\textbf{Ablation Study.}
\
We conduct the ablation study in Table \ref{tb:ab}.
We can find that the proposed ICP and SEI can both help LLMs to reduce hallucinations.
Also, considering only familiarity (i.e., -w/o. Quality RM) ignores other important characteristics of instruction data and limits the instruction-following ability of LLMs.
Thus, even if considering familiarity alone would greatly reduce hallucinations, it is still necessary to introduce a quality reward model to maintain a good balance between following instructions and reducing hallucinations.

\noindent
\textbf{Scalability Study.}
\
We implement our method on the LLaMA-3-70B in Table \ref{tb:ab} to explore whether \OURS~can fit in larger LLMs.
Results indicate that \OURS~scales effectively to larger-scale models.

\begin{table}[t]
    \scriptsize
    \centering
    \resizebox{\linewidth}{!}{
        \begin{tabular}{m{\hsize}}
        \specialrule{1pt}{1pt}{2pt}
        \textbf{Instruction}: 
        Who is the character Geralt of Rivia from The Witcher series? \\
        \specialrule{1pt}{0pt}{1pt}
        \textbf{Vanilla - 100\%}:
        \sethlcolor{tabcolor5}\hl{Geralt of Rivia is a character and the main protagonist of the fantasy novel series "The Witcher"} written by Andrzej Sapkowski (\cmark). 
        \sethlcolor{tabcolor1}\hl{He is a witcher}, \underline{a rare human mutation born with powers} (\xmark) to \sethlcolor{tabcolor4}\hl{fight against the monsters} that \underline{threaten the world of the living} (\xmark). 
        Geralt is known for his intelligence, skill in combat, and his use of magical artifacts, particularly his silver sword (\cmark).  
        Throughout the series, Geralt faces numerous challenges and battles as he journeys across the Continent, \sethlcolor{tabcolor3}\hl{accompanied by his friends and lovers}, in a quest to protect the world and the people he cares about (\cmark). \\

        \specialrule{0pt}{0pt}{1pt}
        \cdashline{1-1}[0.8pt/2pt]
        \specialrule{0pt}{0pt}{2pt}
        \textbf{\OURS - 5\%}: 
        \sethlcolor{tabcolor5}\hl{Geralt of Rivia is a character from The Witcher series}, a fantasy franchise that includes a series of novels, short stories, and video games (\cmark).  \sethlcolor{tabcolor1}\hl{He is a witcher}, \sethlcolor{tabcolor4}\hl{a member of a guild of professional monster hunters}, and his role is to protect the innocent from supernatural threats (\cmark). 
        Geralt is a complex and brooding character, with a strong sense of honor and duty, and a deep understanding of the world around him (\cmark). 
        He is often portrayed as a loner, but \sethlcolor{tabcolor3}\hl{he is also capable of forming close relationships with others}, particularly with those who share his beliefs and values (\cmark). \\
        \specialrule{1pt}{0.5pt}{0pt}   
        \end{tabular}}
\caption{Case study from LongFact-Objects. We highlight the statements that share the same semantics using the same color.
Models are trained on Alpaca-GPT4.}
    
\label{tab:case}
\end{table}

\begin{table}
\scriptsize	
\centering
\resizebox{0.87\linewidth}{!}{
\begin{tabular}{lcc}
\toprule
\textbf{Model} & \textbf{BioGEN} & \textbf{MT-Bench}\\
\midrule
\rowcolor{blue!5} \textbf{\OURS \ - 5\% - Alpaca-GPT4} & 50.5 & 64.6  \\
\multicolumn{3}{c}{\cellcolor{mypink} \textbf{-w/o ICP}}\\
-w. Confidence Score (Perplexity) & {48.4} & 62.2 \\
-w. Lexical Similarity (Rouge-L) & {47.9} & 61.5 \\
-w. Using Embedding Model & {49.8} & 63.9 \\
\multicolumn{3}{c}{\cellcolor{mypink} \textbf{-w/o SEI}}\\
-w. K-means Clustering via Internal States & {47.8} & 60.2  \\
-w. K-means Clustering via Embedding Model & {48.5} & 63.2 \\
-w Voting without Semantic Clustering & {47.3} & 60.8 \\
\bottomrule
\end{tabular}}
\caption{Evaluation results of \OURS~that employ various methods for measuring the LLM's familiarity.
We report FactScore results on BioGEN.}
\label{tb:var} 
\end{table}

\noindent
\textbf{Case Study.}
\
We conduct a case study in Table \ref{tab:case} to visually show the advantages of \OURS.
Compared to using the whole training data, our method ensures the statements are correct and comprehensive, and the generated text is fluent and natural.

\noindent
\textbf{Variant Methods Testing.}
\
As shown in Table \ref{tb:var}, we further explore the variant methods in measuring the LLM's familiarity.
For ICP, we separately replace it with sentence-level confidence (Perplexity) and lexical metrics (Rouge-L).
Specifically, we use the average perplexity score of generated responses to represent sentence-level confidence and use the average Rouge-L score between each pair of two generated responses as lexical metrics.
However, these straightforward strategies neglect highly concentrated semantic information within the internal states, and thus fail to capture the fine-grained differences between responses and limit the final performance.
We also explore the effectiveness of an advanced embedding model, we use \textsc{text-embedding-3-large}\footnote{https://platform.openai.com/docs/guides/embeddings} from OpenAI and set the dimension as 4096.
We find that using the internal states achieves better performance, showing the effectiveness of our method.
This is because internal states may reflect more dense and fine-grained information from LLM itself that may have been lost in the decoding phase of the responses.
For SEI, we explore whether using k-means clustering based on internal states computed as ICP and sentence embedding from \textsc{text-embedding-3-large} can identify suitable semantic clusters.
We can find that our method achieves better performance because the k-means algorithm is not based on semantic equivalence to get the clusters.
Also, the internal states of LLMs cannot efficiently represent the target response, as this response is manually labeled or generated by other advanced LLMs instead of generated by the LLM itself.
We also find that simply voting based on the textual contents instead of semantic clustering limits the final performance, as these responses are often free-form and can have the same meaning in different ways. 

\noindent
\textbf{Discussion.}
We conduct the parameter study to test the robustness of our method in Appendix \ref{appendix:para}.
We also conduct a transferability study in Appendix \ref{appendix:trans} and find \OURS~can fit in other LLMs.
We further explore the design of our method in Appendix \ref{appendix:design} and find our design is effective.
We conduct a case study in Appendix \ref{appendix:cs-ss} to qualitatively show the difference between samples with different scores.

\begin{figure}
    \centering
    \includegraphics[width=6.4cm]{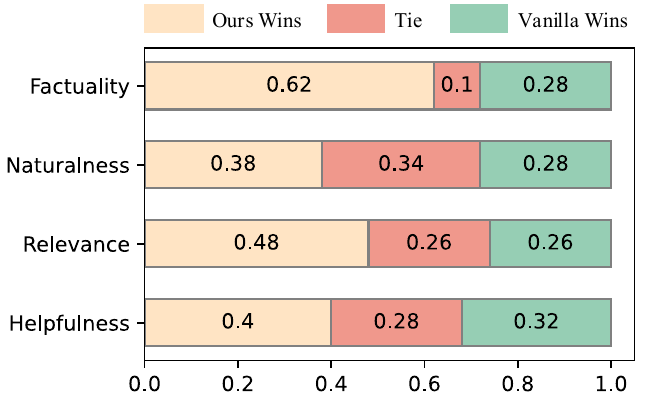}
    \caption{Human evaluation across four key dimensions. The models are trained on Alpaca-GPT4.}
    \label{fig_human}
\end{figure}

\noindent
\textbf{Human Evaluation.}
\
We conduct a human evaluation on the 50 generated biographies from BioGEN across four key dimensions: factuality, helpfulness, relevance, and naturalness.
For each comparison, three options are given (Ours Wins, Tie, and Vanilla Fine-tuning Wins) and the majority voting determines the final result. 
Figure \ref{fig_human} shows that our method significantly reduces hallucinations and effectively follows instructions with high-quality responses.
Details can be found in Appendix \ref{appendix:huamn}.

\section{Conclusion}
\label{subsection:conclusion}
In this paper, we introduce \OURS, a novel framework designed to identify high-quality data that aligns well with the LLM’s learned knowledge to reduce hallucination.
NOVA includes Internal Consistency Probing and Semantic Equivalence Identification, which are designed to separately measure the LLM's familiarity with the given instruction and target response, then prevent the model from being trained on unfamiliar data, thereby reducing hallucinations.
Lastly, we introduce an expert-aligned reward model, considering characteristics beyond just familiarity to enhance data quality.
By considering data quality and avoiding unfamiliar data, we can use the selected data to effectively align LLMs to follow instructions and hallucinate less in the instruction tuning stage.
Experiments and analysis show the effectiveness of \OURS.

\section*{Limitations}
Although empirical experiments have confirmed the effectiveness of the proposed \OURS, two major limitations remain. 
Firstly, our proposed method requires LLMs to generate multiple responses for the given instruction, which introduces additional execution time.
However, it is worth noting that this additional execution time is used to perform offline data filtering, our proposed method does not introduce additional time overhead in the inference phase.
Additionally, \OURS~is primarily used for single-turn instruction data filtering, thus exploring its application in multi-turn scenarios presents an attractive direction for future research.

\section*{Acknowledgements}
We would like to thank the anonymous reviewers for their thoughtful and constructive comments.
This work is supported by the National Science and Technology Major Project (2020AAA0106502), the National Natural Science Foundation of China (No. T2341003), and a grant from the Guoqiang Institute, Tsinghua University.



\bibliography{custom}

\appendix
\section*{Appendix}


\section{Evaluation}
\label{appendix:eva}

In this section, we will detail the benchmarks and evaluation metrics.

\paragraph{BioGEN. (Factuality)}
This benchmark requires generating short biographies for particular people entities, with a total of 500 samples.
The task of generating people biographies is effective, because generations consist of verifiable statements rather than debatable or subjective ones, and the scope is broad (i.e., covering diverse nationalities, professions, and levels of rarity).
To evaluate each generated response, we follow the FactScore procedure to extract the number of correct and incorrect facts.
Following \citet{min2023factscore}, we first employ GPT-3.5-Turbo-0125 to break a generation into a series of atomic facts and utilize GPT-3.5-Turbo-0125 to compute the percentage of atomic facts supported by a reliable knowledge source.
The percentage of the correct statements (\% FactScore), the number of generated statements (\# Facts), and the ratio of generations that do not abstain from responding (\% Respond) are adopted as the evaluation metrics.

\paragraph{LongFact. (Factuality)}
LongFact requests detailed descriptions for a queried entity and expects a document-level response that is typically very long, often exceeding a thousand tokens.
Specifically, LongFact consists of two subtasks: \textbf{LongFact-Concepts} and \textbf{LongFact-Objects}, separated based on whether the questions ask about concepts or objects.
Following \citet{cheng2024integrativedecodingimprovefactuality}, we use 120 samples of each task for evaluation.
The evaluation process is similar to BioGEN.
We employ GPT-3.5-Turbo-0125 and report the FactScore of LongFact-Concepts and LongFact-Objects, termed as \% Concepts and \% Objects.

\paragraph{FollowRAG. (Faithfulness and Instruction Following)}
FollowRAG aims to assess the model’s ability to follow user instructions in complex multi-document contexts, covering 22 fine-grained atomic instructions across 6 categories. 
The queries in FollowRAG are sourced from 4 QA datasets across NaturalQA \citep{47761}, TriviaQA \citep{joshi-etal-2017-triviaqa}, HotpotQA \citep{yang-etal-2018-hotpotqa}, and WebQSP \citep{Yih2016TheVO}.
It collects and verifies definitions and examples of atomic instructions using rules (e.g., code), excluding those irrelevant to retrieval-augmented generation (RAG) scenarios.
FollowRAG identifies 22 types of instruction constraints, encompassing language, length, structure, and keywords.
Thus, it is suitable to use FollowRAG to evaluate the model’s ability to follow user instructions.
Utilizing the verifiable nature of designed atomic instructions, FollowRAG automates the verification of the model’s adherence to each instruction through code validation.
We calculate the average pass rate for each atomic instruction across all samples to determine the instruction-following score and name this task as \textbf{FollowRAG-Intruction}.
Also, FollowRAG provides retrieved passages as contextual information to evaluate the model's faithfulness.
We name this task as \textbf{FollowRAG-Faithfulness}.
Under new instruction constraints, the model’s target output differs from the gold answers in the original QA dataset, rendering traditional metrics like EM ineffective.
Following \citet{dong2024generalinstructionfollowingalignmentretrievalaugmented}, we use the original gold answers as a reference and utilize GPT-4o-2024-05-13 to evaluate whether the model’s outputs address the questions.
The scoring criteria are as follows: Completely correct (1 point), Partially correct (0.5 points), Completely incorrect (0 points). 
The average score of all samples is taken as the final score for FollowRAG-Faithfulness.

\paragraph{MT-Bench. (Instruction Following)}
MT-Bench is a benchmark consisting of 80 questions, designed to test instruction-following ability, covering common use cases and challenging questions. 
It is also carefully constructed to differentiate chatbots based on their core capabilities, including writing, roleplay, extraction, reasoning, math, coding, STEM knowledge, and social science. 
For evaluation, MT-Bench prompts GPT-4 to act as judges and assess the quality of the models’ responses. 
For each turn, GPT-4 will give a score on a scale of 10. 
Notably, since we only fine-tune on single-turn instruction data (e.g., Alpaca and Alpaca-GPT4), the evaluation is restricted to Turn 1 of MTBench, similar to previous studies \citep{li2024shot}.

\section{Implementation Details}
\label{appendix:id}

\paragraph{Hyperparameters and Devices.}
We use Adam optimizer \citep{kingma2017adammethodstochasticoptimization} to train our model, with a $2\times10^{-5}$ learning rate and a batch size of 16, steers the training across three epochs.
We set the maximum input length for the models to 1024. 
To get the generated initial responses for knowledge estimation, we set the temperature as 0.7 and set hyperparameter $K$ as 10 to generate 10 responses for the given instruction $q$.
We conduct our experiments on NVIDIA A800 80G GPUs with DeepSpeed+ZeRO3 and BF16.

\paragraph{Training of NLI Model.}
Natural language inference (NLI) is a well-studied task in the NLP community.
We employ a well-trained NLI model DeBERTa-large-mnli\footnote{https://huggingface.co/microsoft/deberta-large-mnli} \citep{he2021deberta} (0.3B) as our model to conduct the experiments and report the results.
DeBERTa-large-mnli is the DeBERTa large model fine-tuned with multi-genre natural language inference (MNLI) corpus \citep{N18-1101}, which is a crowd-sourced collection of 433k sentence pairs annotated with textual entailment information.
DeBERTa-large-mnli shows advanced performance in various NLI benchmarks e.g., 91.5\% accuracy on MNLI test set.

\paragraph{Traning of Quality Reward Model.}
Our training data is derived from an expert-revised dataset \citep{DBLP:conf/icde/LiuTZZMZ0HZZMZY24}, which consists of 3,751 instruction pairs from Alpaca refined by linguistic experts to enhance fluency, accuracy, and semantic coherence between instructions and responses.
Meanwhile, \citet{DBLP:conf/icde/LiuTZZMZ0HZZMZY24} employs the edit distance metric (i.e., Levenshtein distance) to assess the quality of the original instruction pair and revised instruction pair.
Thus, we can treat this edit distance metric as the target reward value and use the point-wise loss function to train the reward model.
Specifically, following \citet{ge2024clustering}, we concatenate instruction pairs as text inputs and use the given reward value in the dataset as the target outputs.
We use the average pooling strategy and introduce the additional feed-forward layer to transform the hidden states of the model into a scalar.
Then we use Mean Squared Error as the loss function to train the reward model.
We select DeBERTa-large \citep{he2021deberta} (0.3B) as our model.
We use Adam optimizer to train our model, with a $1.5\times10^{-5}$ learning rate and a batch size of 8.
We train our model on a single NVIDIA A800.

\paragraph{Prompt Template.}
We use the prompt template from Alpaca \citep{alpaca}.
We keep the same template in training and inference.

\begin{figure}
    \centering
    \includegraphics[width=7.6cm]{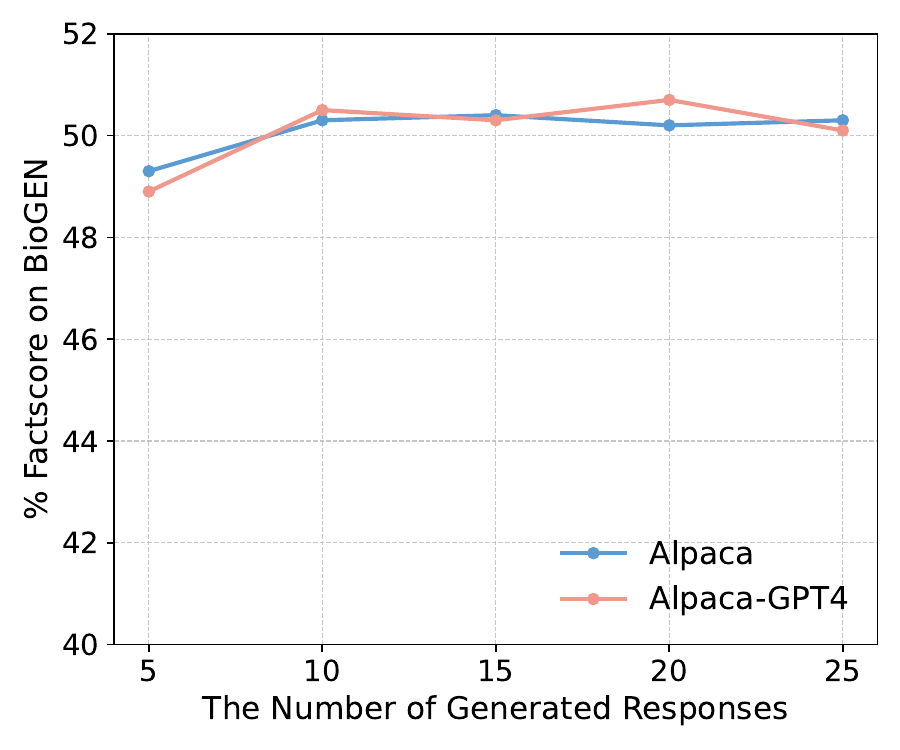}
    \caption{FactScore results on BioGEN with the different number of generated responses $K$. We conduct the experiments based on LLaMA-3-8B. }
    \label{fig_k}
\end{figure}

\section{Parameter Study}
\label{appendix:para}
We explore the effects of two important hyperparameters in our method: the number of generated responses $K$ and the temperature $T$ during the response generation.
As shown in Figure \ref{fig_k}, increasing the number of generated responses improves the performance of our method, but when the number of generated responses is greater than 10, the performance will be stable.
Therefore, we empirically recommend setting the number of generated responses $K$ to 10, which makes our method effective and efficient.
For the temperature $T$, we find that the performance of the model improves as long as the temperature $T$ is chosen wisely and not at an extreme value (e.g., 0, as this would result in multiple generated responses that are exactly the same).
We recommend that the temperature take a moderate value, as this ensures both that there is diversity in the responses generated and that the generated responses do indeed match the model's perceptions (rather than being too random).
Overall, our method \OURS~is robust to these hyperparameters, making our method easy to follow.

\begin{table}
\scriptsize	
\centering
\resizebox{0.75\linewidth}{!}{
\begin{tabular}{lcc}
\toprule
\textbf{Model} & \textbf{Dataset} & \textbf{BioGEN}\\
\midrule
\rowcolor{blue!5} \textbf{\OURS} & Alpaca & \textbf{50.3}   \\
- $T=0$ & Alpaca & 43.2   \\
- $T=0.2$ & Alpaca & 49.3   \\
- $T=0.7$ (Ours) & Alpaca & \textbf{50.3}   \\
- $T=1.0$ & Alpaca & 50.1   \\
- $T=1.3$ & Alpaca & 49.7   \\
\hdashline[2pt/3pt]
\rowcolor{blue!5} \textbf{\OURS } & Alpaca-GPT4 & \textbf{50.5}   \\
- $T=0$ & Alpaca-GPT4 & 43.6   \\
- $T=0.2$ & Alpaca-GPT4 & 48.9   \\
- $T=0.7$ (Ours) & Alpaca-GPT4 & \textbf{50.5}   \\
- $T=1.0$ & Alpaca-GPT4 & 49.8   \\
- $T=1.3$ & Alpaca-GPT4 & 49.5   \\
\bottomrule
\end{tabular}}
\caption{FactScore results on BioGEN with different temperature $T$ during the response generation. 
We conduct the experiments on LLaMA-3-8B and use 5\% selected instruction data from different datasets.}
\label{tb:para} 
\end{table}

\section{Transferability Study}
\label{appendix:trans}
To verify the transferability of the \OURS~method, we conducted experiments on different foundation models using the Alpaca instruction dataset shown in Table \ref{tb:1-hall} and Table \ref{tb:1-if}.
We select LLaMA \citep{llama1} and Qwen-2 \citep{yang2024qwen2technicalreport} at the 7B size as the new base models.
We aim to gain deeper insights into the applicability of the \OURS~method across different models, providing a reference for further research and applications. 
We find that the \OURS~method is also applicable to other models, showing strong transferability and robustness to other models and further research.
Compared to other baselines, \OURS~significantly reduces hallucinations and keeps a strong ability to follow instructions.

\begin{table*}
\scriptsize
\centering  
\resizebox{\textwidth}{!}{
\begin{tabular}{lcccccccccccccc}
\toprule
\multirow{2}{*}{\textbf{Model}} & \multicolumn{3}{c}{\textbf{BioGEN$^\dag$ }} & \multicolumn{3}{c}{\textbf{LongFact$^\dag$}} & \multicolumn{5}{c}{\textbf{FollowRAG - Faithfulness$^\ddag$}} \\
\cmidrule(lr){2-4} \cmidrule(lr){5-7} \cmidrule(lr){8-12} 
& \textbf{FactScore} & \textbf{Respond} & \textbf{Facts}  & \textbf{Objects} & \textbf{Concepts} & \textbf{Avg.} & \textbf{NaturalQA} & \textbf{TriviaQA} & \textbf{HotpotQA} & \textbf{WebQSP} & \textbf{Avg.}\\
\midrule
\multicolumn{12}{c}{\cellcolor{myyellow} \textbf{LLaMA-1}} \\
Vanilla - 100\% & 38.6 & 100.0 & 16.6 & 84.3 &78.2 &  81.3 & 37.5 	& 	 	50.5 &	 	 	16.0 	 &	 	47.5 	 &	 	37.9  \\
FLAME-DPO$^{fact}$ & 41.2 & 100.0 & 14.8 & 86.7 & 81.2  & 84.0  &41.5 & 			55.0 & 			21.5 		& 	52.5 	& 		42.6 \\
SELF-EVAL & 41.8 & 100.0 & 15.7 & 87.0 & 80.8 & 83.9  &42.5 	 &		56.5 		 &	22.5 		 &	53.5 		 &	43.8 \\
\midrule
IFD - 5\% & 40.2 & 100.0 & 20.1 & 83.2 & 80.4 & 81.8 &38.0 		&	53.5 		&	18.5 	&		49.0 		&	39.8 \\
CaR - 5\% & 39.6 & 100.0 & 18.2  & 85.9 & 80.1 & 83.0 &38.0 		&	53.0 	&		19.0 	&		50.5 		&	40.1 \\
Nuggets - 5\% & 39.3 & 100.0 & 19.4 & 85.1 & 77.3 & 81.2 &39.5 	&		54.5 	&		20.0 		&	50.0 	&		41.0 \\
\rowcolor{blue!5} \textbf{NOVA - 5\%} & \textbf{43.6} & 100.0 & \textbf{21.5} & \textbf{88.1} & \textbf{82.5} & \textbf{85.3}  &\textbf{44.5} 		&	\textbf{58.5} 	&		\textbf{24.0} 	&		\textbf{55.5} 	&		\textbf{45.6} \\
\hdashline[2pt/3pt]
\rowcolor{blue!5} $\Delta$ compared to Vanilla - 100\% & \textcolor[rgb]{0.7,0,0}{+5.0} & - & \textcolor[rgb]{0.7,0,0}{+4.9}  & \textcolor[rgb]{0.7,0,0}{+3.8} & \textcolor[rgb]{0.7,0,0}{+4.3} & \textcolor[rgb]{0.7,0,0}{+4.1} & \textcolor[rgb]{0.7,0,0}{+7.0} 	&		\textcolor[rgb]{0.7,0,0}{+8.0} &			\textcolor[rgb]{0.7,0,0}{+8.0} 	&		\textcolor[rgb]{0.7,0,0}{+8.0} 	&		\textcolor[rgb]{0.7,0,0}{+7.7} \\
\midrule
IFD - 10\% & 40.7 & 100.0 & 19.2 & 85.2 & 80.3  & 82.8  &40.0 	&		54.5 		&	20.0 		&	51.0 	&		41.4  \\
CaR - 10\% & 40.3 & 100.0 & \textbf{21.1} & 83.4 & 79.2  & 81.3& 41.0 	&		52.0 	&		18.0 	&		49.5 		&	40.1 \\
Nuggets - 10\% & 41.0 & 100.0 & 18.8  & 84.2 & 78.6 & 81.4 & 39.5 		& 	53.0 		& 	17.5 		& 	51.0 		& 	40.3 \\
\rowcolor{blue!5} \textbf{NOVA - 10\%} & \textbf{43.2} & 100.0 & 20.7  & \textbf{87.6} & \textbf{83.2} & \textbf{85.4}& \textbf{43.5} 	&		\textbf{59.5} 			& \textbf{22.5} 		&	\textbf{53.0} 	&		\textbf{44.6}  \\
\hdashline[2pt/3pt]
\rowcolor{blue!5} $\Delta$ compared to Vanilla - 100\% & \textcolor[rgb]{0.7,0,0}{+4.6} & - & \textcolor[rgb]{0.7,0,0}{+4.1} & \textcolor[rgb]{0.7,0,0}{+3.3} & \textcolor[rgb]{0.7,0,0}{+5.0}  & \textcolor[rgb]{0.7,0,0}{+4.2} & \textcolor[rgb]{0.7,0,0}{+6.0} 	&		\textcolor[rgb]{0.7,0,0}{+9.0} 		&	\textcolor[rgb]{0.7,0,0}{+6.5} 	&		\textcolor[rgb]{0.7,0,0}{+5.5} 	&		\textcolor[rgb]{0.7,0,0}{+6.7} \\
\midrule
IFD - 15\% & 39.2 & 100.0 & 18.7 & 86.1 & 81.1  & 83.6 & 39.5 	&		52.0 		&	17.5 	&		49.5 	&		39.6 \\
CaR - 15\% & 40.2 & 100.0 & 19.3 & 84.2 & 80.4  & 82.3 & 38.0 	&		51.5 	&		17.0 		&	48.0 		&	38.6 \\
Nuggets - 15\% & 40.9 & 100.0 & 18.1 & 83.3  & 80.0 & 81.7 &40.0 		&	52.5 		&	15.5 		&	50.5 &			39.6 \\
\rowcolor{blue!5} \textbf{NOVA - 15\%} & \textbf{44.1} & 100.0 & \textbf{19.4} & \textbf{89.6} & \textbf{83.7}  & \textbf{86.7} &\textbf{42.5} 		&	\textbf{56.5} 		&	\textbf{23.5} 	&		\textbf{54.5} 	&		\textbf{44.3}   \\
\hdashline[2pt/3pt]
\rowcolor{blue!5} $\Delta$ compared to Vanilla - 100\% & \textcolor[rgb]{0.7,0,0}{+5.5} & - & \textcolor[rgb]{0.7,0,0}{+2.8}  & \textcolor[rgb]{0.7,0,0}{+5.3} & \textcolor[rgb]{0.7,0,0}{+5.5} & \textcolor[rgb]{0.7,0,0}{+5.4} &\textcolor[rgb]{0.7,0,0}{+5.0} 	&		\textcolor[rgb]{0.7,0,0}{+6.0} 	&		\textcolor[rgb]{0.7,0,0}{+7.5} &			\textcolor[rgb]{0.7,0,0}{+7.0} 		&	\textcolor[rgb]{0.7,0,0}{+6.4} \\
\midrule
\multicolumn{12}{c}{\cellcolor{myyellow} \textbf{Qwen-2}} \\
Vanilla - 100\% & 40.3 & 100.0 & 17.3 &83.4&	80.2&	81.8 	 & 39.5 & 57.5 & 18.5 & 49.0 & 41.1  \\
FLAME-DPO$^{fact}$ & 47.1	&100.0	&16.9	&87.8&	82.7&	85.3 	& 44.5 & 58.0 & 20.5 & 53.0 & 44.0  \\
SELF-EVAL & 46.8	&100.0 	&14.2	&88.2&	81.6&	84.9  & 43.5 & 59.0 & 21.0 & 53.0 & 44.1 \\
\midrule
IFD - 5\% &44.2&	100.0 	&16.5&	85.2&	81.2	&83.2  & 42.5 & 56.5 & 20.5 & 53.5 & 43.3\\
CaR - 5\% &45.7	&100.0 &	\textbf{18.6}	&84.1	&81.5	&82.8 & 44.5 & 55.5 & 21.0 & 52.0 & 43.3 \\
Nuggets - 5\% & 46.6&	100.0 	&17.8	&84.7&	81.0 &	82.9 & 43.0 & 57.5 & 21.5 & 52.5 & 43.6 \\
\rowcolor{blue!5} \textbf{NOVA - 5\%} & \textbf{49.1}	&100.0& 	18.3 &	\textbf{90.2}&	\textbf{83.2}	& \textbf{86.7} & \textbf{46.0} & \textbf{59.6} & \textbf{23.5} & \textbf{55.5} & \textbf{46.1} \\
\hdashline[2pt/3pt]
\rowcolor{blue!5} $\Delta$ compared to Vanilla - 100\% & \textcolor[rgb]{0.7,0,0}{+8.8} &	-	& \textcolor[rgb]{0.7,0,0}{+1.0} 	& \textcolor[rgb]{0.7,0,0}{+6.8} 	& \textcolor[rgb]{0.7,0,0}{+3.0} 	& \textcolor[rgb]{0.7,0,0}{+4.9}  & \textcolor[rgb]{0.7,0,0}{+6.5} & \textcolor[rgb]{0.7,0,0}{+2.1} & \textcolor[rgb]{0.7,0,0}{+5.0} & \textcolor[rgb]{0.7,0,0}{+6.5} & \textcolor[rgb]{0.7,0,0}{+5.0}\\
\midrule
IFD - 10\% & 44.5	& 100.0 &	17.8&	84.2	&80.5&	82.4  & 41.5 & 59.5 & 19.5 & 51.0 & 42.9 \\
CaR - 10\%& 45.2&	100.0 &	20.3&	84.5	&79.8&	82.2  & 42.5 & 60.0 & 18.5 & 53.0 & 43.5 \\
Nuggets - 10\% & 46.1	&100.0& 	\textbf{23.5}&	85.2&	79.7	& 82.5  & 42.0 & 60.0 & 20.0 & 51.5 & 43.4 \\
\rowcolor{blue!5} \textbf{NOVA - 10\%} & \textbf{47.5}	& \textbf{100.0} 	 & 18.6  &	\textbf{89.6}	 & \textbf{83.5} &	\textbf{86.6}  & \textbf{45.0} & \textbf{62.0} & \textbf{21.5} & \textbf{53.5} & \textbf{45.5} \\
\hdashline[2pt/3pt]
\rowcolor{blue!5} $\Delta$ compared to Vanilla - 100\% & \textcolor[rgb]{0.7,0,0}{+7.2} &	-	& \textcolor[rgb]{0.7,0,0}{+1.3} 	& \textcolor[rgb]{0.7,0,0}{+6.2} 	& \textcolor[rgb]{0.7,0,0}{+3.3} 	& \textcolor[rgb]{0.7,0,0}{+4.7} & \textcolor[rgb]{0.7,0,0}{+5.5} & \textcolor[rgb]{0.7,0,0}{+4.5} & \textcolor[rgb]{0.7,0,0}{+3.0} & \textcolor[rgb]{0.7,0,0}{+4.5} & \textcolor[rgb]{0.7,0,0}{+4.4} \\
\midrule
IFD - 15\% & 43.7	 &100.0 &	19.2	&82.5	&79.5	&81.0  & 42.0 & 61.5 & 18.5 & 52.0 & 43.5 \\
CaR - 15\% & 44.8&	100.0 	&20.8&	81.2	&81.3	&81.3 & 43.0 & 62.5 & 19.5 & 53.0 & 44.5\\
Nuggets - 15\% & 45.7&	100.0 &	\textbf{21.7} &	80.8	&80.1	&80.5 & 40.5 & 62.5 & 20.0 & 52.5 & 43.9\\
\rowcolor{blue!5} \textbf{NOVA - 15\%} & \textbf{47.2} &	100.0 	& 19.3 &	\textbf{88.8}	& \textbf{82.9} &	\textbf{85.9} & \textbf{44.5} & \textbf{64.5} & \textbf{22.0} & \textbf{54.0} & \textbf{46.3}  \\
\hdashline[2pt/3pt]
\rowcolor{blue!5} $\Delta$ compared to Vanilla - 100\% & \textcolor[rgb]{0.7,0,0}{+6.9} &	-&	\textcolor[rgb]{0.7,0,0}{+2.0} 	& \textcolor[rgb]{0.7,0,0}{+5.4} &	\textcolor[rgb]{0.7,0,0}{+2.7} 	& \textcolor[rgb]{0.7,0,0}{+4.0}  & \textcolor[rgb]{0.7,0,0}{+5.0} & \textcolor[rgb]{0.7,0,0}{+5.0} & \textcolor[rgb]{0.7,0,0}{+3.5} & \textcolor[rgb]{0.7,0,0}{+5.0} & \textcolor[rgb]{0.7,0,0}{+5.2} \\
\bottomrule
\end{tabular}
}
\caption{Results on three hallucination benchmarks. 
$\dag$ indicates the factuality hallucination benchmark. $\ddag$ indicates the faithfulness hallucination benchmark. 
We conduct the experiments based on Alpaca dataset.}
\label{tb:1-hall}
\end{table*}

\begin{table}[t]
\scriptsize	
\centering
\resizebox{\linewidth}{!}{
\begin{tabular}{lcc}
\toprule
\textbf{Model} & \textbf{MT-Bench} & \textbf{FollowRAG-Intruction}\\
\midrule
\multicolumn{3}{c}{\cellcolor{myyellow} \textbf{LLaMA-1}}\\
Vanilla - 100\% & 47.8 & 37.7  \\
FLAME-DPO$^{fact}$ & 40.6 & 37.5 \\
SELF-EVAL & 42.2 & 38.1 \\
\midrule
IFD - 5\% & 48.3 & 37.8 \\
CaR - 5\% & \textbf{50.1} & \textbf{38.2} \\
Nuggets - 5\% & 48.6 & 38.0 \\
\rowcolor{blue!5} \textbf{NOVA - 5\%} & 49.8 & 38.1 \\
\hdashline[2pt/3pt]
\rowcolor{blue!5} $\Delta$ compared to Vanilla - 100\% & \textcolor[rgb]{0.7,0,0}{+2.0} & \textcolor[rgb]{0.7,0,0}{+0.4} \\
\midrule
IFD - 10\% & 47.9 & 38.6 \\
CaR - 10\% & \textbf{49.5} & 38.1 \\
Nuggets - 10\% & 48.4 & 38.7 \\
\rowcolor{blue!5} \textbf{NOVA - 10\%} & 49.3 & \textbf{39.0} \\
\hdashline[2pt/3pt]
\rowcolor{blue!5} $\Delta$ compared to Vanilla - 100\% & \textcolor[rgb]{0.7,0,0}{+1.5} & \textcolor[rgb]{0.7,0,0}{+1.3} \\
\midrule
IFD - 15\% & 48.5 & 38.2 \\
CaR - 15\% & \textbf{50.3} & 37.6 \\
Nuggets - 15\% & 49.5 & \textbf{38.6} \\
\rowcolor{blue!5} \textbf{NOVA - 15\%} & 48.3 & 38.0 \\
\hdashline[2pt/3pt]
\rowcolor{blue!5} $\Delta$ compared to Vanilla - 100\% & \textcolor[rgb]{0.7,0,0}{+0.5} & \textcolor[rgb]{0.7,0,0}{+0.4} \\
\midrule
\multicolumn{3}{c}{\cellcolor{myyellow} \textbf{Qwen-2}}\\
Vanilla - 100\% & 50.2 & 38.2 \\
FLAME-DPO$^{fact}$ & 47.8 & 38.7 \\
SELF-EVAL & 49.5 & 37.3 \\
\midrule
IFD - 5\% & 59.5 & 39.2 \\
CaR - 5\% & \textbf{61.2} & 39.5 \\ 
Nuggets - 5\% & 60.3 & \textbf{40.2} \\
\rowcolor{blue!5} \textbf{NOVA - 5\%} & 60.8 & 39.7  \\
\hdashline[2pt/3pt]
\rowcolor{blue!5} $\Delta$ compared to Vanilla - 100\% & \textcolor[rgb]{0.7,0,0}{+10.6} & \textcolor[rgb]{0.7,0,0}{+1.5} \\
\midrule
IFD - 10\% & 59.8 & 40.1 \\
CaR - 10\% & \textbf{60.1} & 40.5 \\
Nuggets - 10\% & 58.8 & \textbf{41.1} \\
\rowcolor{blue!5} \textbf{NOVA - 10\%} & 58.4 & 40.1 \\
\hdashline[2pt/3pt]
\rowcolor{blue!5} $\Delta$ compared to Vanilla - 100\% & \textcolor[rgb]{0.7,0,0}{+8.2} & \textcolor[rgb]{0.7,0,0}{+1.9} \\
\midrule
IFD - 15\% & \textbf{59.3} & \textbf{40.5} \\
CaR - 15\% & 57.5 & 39.8 \\
Nuggets - 15\% & 58.5 & 40.3 \\
\rowcolor{blue!5} \textbf{NOVA - 15\%} & 59.2 & 40.0 \\
\hdashline[2pt/3pt]
\rowcolor{blue!5} $\Delta$ compared to Vanilla - 100\% & \textcolor[rgb]{0.7,0,0}{+9.0} & \textcolor[rgb]{0.7,0,0}{+1.8} \\
\bottomrule
\end{tabular}}
\caption{Results on two instruction-following benchmarks based on Alpaca dataset.}
\label{tb:1-if} 
\end{table}

\section{Design Exploration}
\label{appendix:design}

\textbf{The Design of NLI Model}
\
We further explore the effects of the NLI model on the final performance of \OURS.
We first attempt to analyze the effect of the size of the model on the final results.
Specifically, we introduce DeBERTa-base-mnli\footnote{https://huggingface.co/microsoft/deberta-base-mnli}, DeBERTa-xlarge-mnli\footnote{https://huggingface.co/microsoft/deberta-xlarge-mnli} and DeBERTA-xxlarge-mnli\footnote{https://huggingface.co/microsoft/deberta-v2-xxlarge-mnli}.
As shown in Table \ref{tb:nli_size}, we can find that increasing the size of the NLI model can provide some improvement in the final result, especially when changing the DeBERTa-base-mnli to DeBERTa-large-mnli.
However, continuing to increase the model parameters did not have a significant impact on the final performance. 
Therefore, in order to balance the performance and the inference time of NLI models, we select the DeBERTa-large-mnli to report the final results in our paper.
Meanwhile, we further explore whether we use the advanced LLMs (e.g., GPT-4o and GPT-3.5-Turbo) to directly identify the semantic equivalence and get the correct semantic clusters.
Specifically, we use the prompt shown in Figure \ref{fig:prmpt} to test the generated responses and the target response by querying the advanced LLMs to identify semantic equivalence.
We use the same method as SEI, utilizing the outputs of advanced LLMs to derive semantic clusters and calculate the score of $F_{res}(r)$.
As shown in Table \ref{tb:nli_size}, the direct application of results from advanced LLMs proves effective in identifying semantic equivalence.
Nevertheless, using NLI models delivers competitive or superior final performance while avoiding API-related costs.
Consequently, employing NLI models to identify semantic equivalence is both efficient and effective, substantiating the efficacy of our designed SEI approach.

\begin{table}
\scriptsize	
\centering
\resizebox{0.72\linewidth}{!}{
\begin{tabular}{lcc}
\toprule
\textbf{Model} & \textbf{Size} & \textbf{BioGEN}\\
\midrule
\multicolumn{3}{c}{\cellcolor{myyellow} Alpaca} \\
DeBERTa-base-mnli & 0.1B & 49.7   \\
DeBERTa-large-mnli & 0.3B & 50.3   \\
DeBERTa-xlarge-mnli & 0.7B & 50.1   \\
DeBERTa-xxlarge-mnli & 1.3B & \textbf{50.5}   \\
\hdashline[2pt/3pt]
GPT-3.5-Turbo-0125 & unknown & 49.8 \\
GPT-4o-2024-05-13 & unknown & 50.2 \\
\midrule
\multicolumn{3}{c}{\cellcolor{myyellow} Alpaca- GPT4} \\
DeBERTa-base-mnli & 0.1B & 49.4   \\
DeBERTa-large-mnli & 0.3B & 50.5   \\
DeBERTa-xlarge-mnli & 0.7B & \textbf{51.2}   \\
DeBERTa-xxlarge-mnli & 1.3B & 50.3   \\
\hdashline[2pt/3pt]
GPT-3.5-Turbo-0125 & unknown & 49.2 \\
GPT-4o-2024-05-13 & unknown & 50.0 \\
\bottomrule
\end{tabular}}
\caption{FactScore results on BioGEN with different models. We conduct experiments on LLaMA-3-8B and use selected 5\% data from different datasets.}
\label{tb:nli_size} 
\end{table}

\begin{table}
\scriptsize	
\centering
\resizebox{0.7\linewidth}{!}{
\begin{tabular}{lc}
\toprule
\textbf{Model}  & \textbf{Accuracy}\\
\midrule
\rowcolor{blue!5} \textbf{Our Used Reward Model} &  \textbf{92.0}  \\
GPT-3.5-Turbo-0125  & 85.0 \\
GPT-4o-2024-05-13  & 90.0 \\
\bottomrule
\end{tabular}}
\caption{Accuracy of our used reward model and other advanced LLMs on the constructed test set.}
\label{tb:rm} 
\end{table}

\noindent
\textbf{The Design of Quality Reward Model}
\
We also explore the effectiveness of the quality reward model.
We introduce UltraFeedback \citep{cui2024ultrafeedback} and sample 100 instructions and their corresponding responses as the test set (we find that most of the selected data are in English, but some of the selected instruction types are translation tasks, so a few data contain Chinese responses).
Specifically, for each instruction, we randomly select 2 responses and determine the ranking between the responses based on their labeled scores of instruction-following, honesty, truthfulness, and helpfulness.
Only if all four scores are higher will the response be considered a high-quality response.
Meanwhile, we involve two Ph.D. students to conduct the human evaluation to ensure the correctness of the response ranking of each sample.
Afterwards, we take the instructions and the responses as inputs to each model, and let the model determine the ranking between the responses and calculate the accuracy of the model's prediction of the ranking.
We compare our used Quality Reward Model with GPT-3.5-Turbo-0125 and GPT-4o-2024-05-13.
We use the same prompt for each model as \citet{ge2024clustering}.
As shown in Table \ref{tb:rm}, our reward model achieves better performance, showing the effectiveness of our method.
Despite GPT-4o’s strong alignment with human preferences in most general tasks, our reward model trained on the expert-revised preference dataset can perform better, highlighting the subtle gap between expert preferences and advanced GPT-4o preferences.

\noindent
\textbf{The Design of Obtaining Sentence Embedding. Alpaca-GPT4}
\
For $K$ generated responses, we use the internal states of the last token of each response in the last layer as the final sentence embeddings $E=[e_1,e_2,...,e_K]$, as it effectively captures the sentence semantics \citep{azaria2023the}.
We further explore the different ways to obtain sentence embedding.
Specifically, we first average all the internal states of tokens in the sentence to obtain the sentence embedding (named Average Pooling), which is an intuitive method to get the sentence embedding for decoder-only models.
As shown in Table \ref{tb:sentence}, we can find the design of \OURS~achieves better performance in both reducing hallucinations and following instructions, showing the effectiveness of our designed SEI.
We further explore the internal states from which layer in the LLMs can be used to effectively measure the consistency.
Except for the internal states from the last layer, we select both internal states from the first layer and internal states from the middle layer (layer 16 for LLaMA-3-8B), and use the internal states of the last token to represent the sentence embeddings.
We can find that using sentence embedding in the shallow layer yields inferior performance compared to using sentence embedding in the deep layers, as the shallow layer may not effectively model the rich semantic information.
Overall, extensive experiments show that our design of \OURS~is sound and effective.

\begin{figure*}[t]
    \centering
    \begin{tcolorbox}[title = {The Prompt for Identifying the Semantic Equivalence}, size=title, colframe = white, colbacktitle = black!65!white]
    \noindent   
    Please compare the following two sentences and determine whether they are semantically the same. If they are semantically identical, respond with "Identical"; if not, respond with "Different." Consider the meaning, context, and any implicit nuances of the sentences. \\

    Sentence 1: \{Sentence 1\} \\
    Sentence 2: \{Sentence 1\} \\
    
    Provide your judgment below:
    \end{tcolorbox}
    \caption{The prompt for identifying the semantic equivalence.}
    \label{fig:prmpt}
\end{figure*}

\begin{table}
\scriptsize	
\centering
\resizebox{0.85\linewidth}{!}{
\begin{tabular}{lcc}
\toprule
\textbf{Model} & \textbf{BioGEN} & \textbf{MT-Bench}\\
\midrule
\rowcolor{blue!5} \textbf{\OURS \ - 5\%} & \textbf{50.5} & \textbf{64.6}  \\
-w. Average Pooling & 49.5 & 64.2 \\
-w. The First Layer & 48.9 & 63.7 \\
-w. The Middle Layer & 49.8 & 64.4 \\
-w. The Last Layer (Ours) & 50.5 & 64.6 \\
\bottomrule
\end{tabular}}
\caption{Evaluation results of \OURS~that employ various methods for obtaining sentence embedding. We conduct the experiments based on LLaMA-3-8B and the Alpaca-GPT4 dataset.
We report the FactScore results on BioGEN.}
\label{tb:sentence} 
\end{table}

\begin{table}
\scriptsize	
\centering
\resizebox{1\linewidth}{!}{
\begin{tabular}{lcc}
\toprule
\textbf{Model} & \textbf{BioGEN} & \textbf{MT-Bench}\\
\midrule
\rowcolor{blue!5} \textbf{\OURS~- LLaMA-3-8B - 5\%} & \textbf{50.3} & \textbf{60.5}  \\
-w/o. Few-shot Demonstrations & 50.1 & 59.8 \\
\rowcolor{blue!5} \textbf{\OURS~- LLaMA-1-7B - 5\%} & \textbf{43.6} & \textbf{49.8}  \\
-w/o. Few-shot Demonstrations & 41.9 & 49.2 \\
\bottomrule
\end{tabular}}
\caption{The effects of used few-shot demonstrations. 
We conduct the experiments based on two base models and the Alpaca dataset.
We report the FactScore results on BioGEN.}
\label{tb:de} 
\end{table}

\noindent
\textbf{The Design of Using Few-shot Demonstration.}
\
As detailed in Sec. \ref{section:IKP}, we sample $K$ responses $[r'_1,...,r'_K]$ from a base LLM with few-shot demonstrations \citep{lin2024the} to ensure the coherence of generated responses.
We use the same demonstrations as \citet{lin2024the}.
We further conduct experiments to explore the effects of these used demonstrations.
We find that using few-shot demonstrations in the process of generating responses for a given instruction allows the base LLMs to better express what they have learned in the pre-training stage.
In turn, this will enable ICP and SEI to better estimate the knowledge contained in the instruction data and thus better identify the high-quality instruction data that aligns well with the LLM’s learned knowledge to reduce hallucination and improve instruction-following ability.
At the same time, we find that this strategy improves more for base models with poor capabilities (e.g., LLaMA-1-7B), which is due to the fact that a poor base LLM may hold relevant knowledge in response to a query, yet occasionally falters in conveying accurate information \citep{zhang-etal-2024-self}.

\section{Human Evaluation}
\label{appendix:huamn}
During the human evaluation, the participants follow the principles in Figure \ref{fig:human_evaluation_principles} to make the decision.
For each comparison, three options are given (Ours Wins, Tie, and Vanilla Fine-tuning Wins) and the majority voting determines the final result. 
We invite three Ph.D. students to compare the responses generated by the models.
Before participants begin to make judgments, we describe the principles of our design in detail and ensure that each participant correctly understands the principles.
If the final result can not be determined by the majority voting, we will make the discussion among the participants and vote on the result again.

\begin{figure*}[t]
    \centering
    \begin{tcolorbox}[title = {The Principles of Human Evaluation}, size=title, colframe = white, colbacktitle = black!65!white]
    \noindent   
    You are asked to evaluate the biographies generated by different models.
    You should choose the preferred biography according to the following perspectives independently: \\

    1. \textbf{Factuality}: Whether the biography provides relatively more factual statements over the non-factual statements? \\
    2. \textbf{Helpfulness}: Whether the biography provides useful information? \\
    3. \textbf{Relevance}: Whether the statements contained in the biography relevant to the provided people entity? \\
    4. \textbf{Naturalness}: Whether the biography sound natural and fluent? \\

    Finally, please make a decision among 3 opinions, including Win, Tie, and Loss.

    \end{tcolorbox}
    \caption{The principles of human evaluation.}
    \label{fig:human_evaluation_principles}
\end{figure*}

\section{Case Study for Selected Samples}
\label{appendix:cs-ss}
To evaluate our proposed \OURS~qualitatively, we also select some instruction samples from the Alpaca dataset for case studies as shown in Figure \ref{fig:casestudy}.
Firstly, we can find that simply using $R_{familiarity}$ in Eq. (\ref{eq:final_r}) can effectively identify the simple and straightforward instruction samples that align well with LLM’s knowledge.
On the contrary, the sample ranked last according to $R_{familiarity}$ contains the open-ended instruction and the very subjective target response.
Meanwhile, further using $R_{final}$ in Eq. (\ref{eq:final_r}) that considers characteristics beyond just familiarity $R_{familiarity}$ by introducing the quality reward model further enhances data quality, e.g., the complexity of instructions.
We can also observe that the sample ranked last according to $R_{final}$ also contains the open-ended instruction and the detailed target response, which may introduce familiar knowledge to LLMs and further promote hallucinations.

\newpage
\begin{figure*} 
    \centering
    \begin{tcolorbox}[title = {Case Study for Selected Samples}, size=title, colframe = white, colbacktitle = black!65!white]
    \noindent
    \

    \faLaughBeam~\textbf{Sample ranked first according to $R_{familiarity}$ in Eq. (\ref{eq:final_r}):}
    
    \textbf{Instruction:} Find a word that is opposite in meaning to the given word: assured.
    
    \textbf{Target Response:} Doubtful. \\ \\

    \faFrown~\textbf{Sample ranked last according to $R_{familiarity}$ in Eq. (\ref{eq:final_r}):}
    
    \textbf{Instruction:} Could you pick a set of five songs which will fit our theme: a summer night's party?
    
    \textbf{Target Response:} \\ 1. Billie Eilish – “Bad Guy” \\ 2. Ed Sheeran – “Shape of You” \\ 3. Maroon 5 – “Summer Nights” \\ 4. DNCE – “Cake by the Ocean” \\ 5. Drake – “In My Feelings”. \\ \\
    
    \faLaughBeam~\textbf{Sample ranked first according to $R_{final}$ in Eq. (\ref{eq:final_r}):}
    
    \textbf{Instruction:} Given a dictionary, create a JSON representation. dict = \{"name": "John", "age": 18, "is\_adult": True\}
    
    \textbf{Target Response:} \\ \{\\ \quad "name": "John", \\ "age": 18, \\ \quad "is\_adult": true \\ \} \\ \\

    \faFrown~\textbf{Sample ranked last according to $R_{final}$ in Eq. (\ref{eq:final_r}): }
    
    \textbf{Instruction:} Describe how to impress a customer.
    
    \textbf{Target Response:} Listen to the customer carefully, acknowledge their needs and desires, make sure to follow up on commitments, and strive to exceed expectations. \\ \\

    \end{tcolorbox}
    \caption{Case study for selected samples.}
    \label{fig:casestudy}
\end{figure*}

\end{document}